\documentclass[10pt,twocolumn,letterpaper]{article}

\usepackage{cvpr}              

\usepackage{graphicx}
\usepackage{amsmath}
\usepackage{amssymb}
\usepackage{booktabs}
\usepackage{appendix}

\usepackage[pagebackref,breaklinks,colorlinks]{hyperref}

\usepackage[capitalize]{cleveref}
\crefname{section}{Sec.}{Secs.}
\Crefname{section}{Section}{Sections}
\Crefname{table}{Table}{Tables}
\crefname{table}{Tab.}{Tabs.}

\def\cvprPaperID{*****} 
\def\confName{CVPR}
\def\confYear{2023}

\begin{document}

\title{Aria Everyday Activities Dataset}

\author{Zhaoyang Lv,
Nicholas Charron,
Pierre Moulon,
Alexander Gamino,
Cheng Peng, 
Chris Sweeney, 
\\
Edward Miller,
Huixuan Tang,
Jeff Meissner,
Jing Dong, 
Kiran Somasundaram,
Luis Pesqueira,
\\
Mark Schwesinger,
Omkar Parkhi, 
Qiao Gu,
Renzo De Nardi,
Shangyi Cheng,
Steve Saarinen, 
\\
Vijay Baiyya, 
Yuyang Zou,
Richard Newcombe,
Jakob Julian Engel,
Xiaqing Pan,
Carl Ren
\\
Reality Labs Research, Meta
}

\maketitle

\newcommand{\todo}[1]{\textcolor{red}{Todo:{#1}}}

\begin{figure*}
    \centering
    \includegraphics[width=\linewidth]{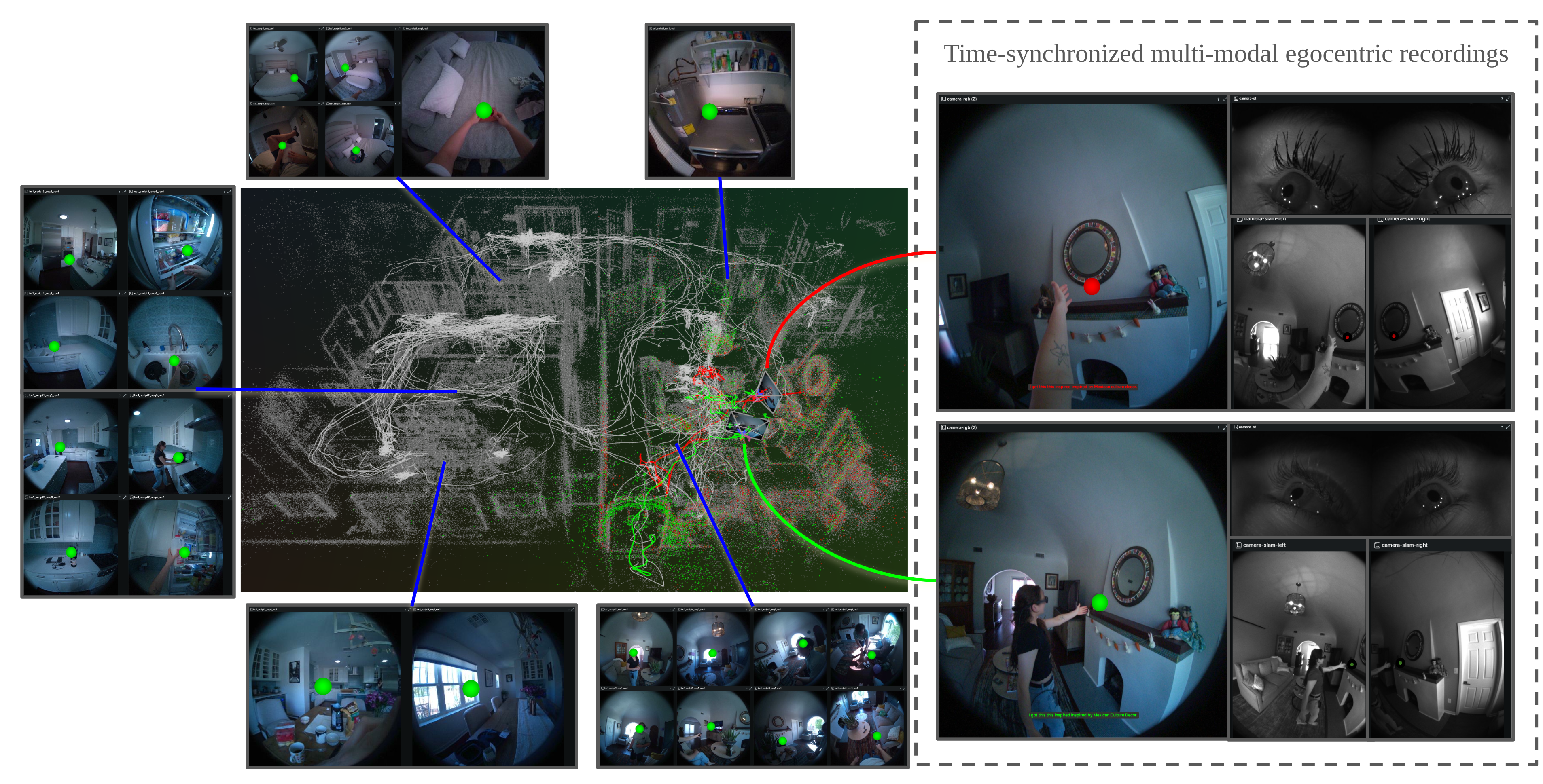}
    \caption{An overview of Aria Everyday Activities (AEA) dataset using some exemplar activities recorded in Location 1. On the right column, we highlight a time-synchronized snapshot of two wearers talking to each other in one activity, with the following information representing each of the viewer in \textcolor{red}{red} and \textcolor{green}{green}: (1) their high-frequency 6DoF close-loop trajectories, (2) observed point cloud, (3) RGB camera view frustum, (4) monochrome scene cameras, (5) eyetracking cameras, (6) their projected eye gaze on all three camera streams and (7) transcribed speech. On the left side, we also highlight a diverse set of activities (e.g. dining, doing laundry, folding clothes, cooking) with the projected eyetracking (\textcolor{green}{green dot}) on the RGB streams. All of the recordings contain close-loop trajectories (white lines) spatially aligned on the environment point cloud (semi-dense points).}
    \label{fig:teaser}
\end{figure*}

\begin{abstract}
We present Aria Everyday Activities (AEA) Dataset, an egocentric multimodal open dataset recorded using Project Aria glasses. AEA contains 143 daily activity sequences recorded by multiple wearers in five geographically diverse indoor locations. Each of the recording contains multimodal sensor data recorded through the Project Aria glasses. In addition, AEA provides machine perception data including high frequency globally aligned 3D trajectories, scene point cloud, per-frame 3D eye gaze vector and time aligned speech transcription. In this paper, we demonstrate a few exemplar research applications enabled by this dataset, including neural scene reconstruction and prompted segmentation. AEA is an open source dataset that can be downloaded from \href{https://www.projectaria.com/datasets/aea/}{projectaria.com}. We are also providing open-source implementations and examples of how to use the dataset in \href{https://github.com/facebookresearch/projectaria_tools}{Project Aria Tools}.

\end{abstract}



\section{Introduction}
\label{sec:intro}

The promise that augmented reality (AR) devices and personal wearable AI devices will be ubiquitous in the future creates the opportunity to develop new technologies that will have profound impacts on people’s lives. New AR \& AI devices that capture always-on multimodal data from the same egocentric point of view as the wearer provide distinctly new data opportunities and challenges. By leveraging these egocentric devices' continuous contextual data, along with advances in machine learning such as large language models, we will be able to build truely personalized and contextualized AI assistants that can act as an extension to the wearer's mind. 

Today's multimodal AI assistants demonstrate intelligent capabilities by leveraging context from texts interleaved with images or short videos, e.g. GPT-4v\cite{gpt4}, Gemini\cite{gemini}, Llava\cite{llava}., but they only have access to public internet data, plus any prompts the users consciously inputs. We believe that future of AI assistants should be able to access significantly more contextual data about the users, and be able to reason with this data. Firstly, the AI should be able to sufficiently leverage all rich sensor data about its wearer and environment using the sensors and hardware we expect in the future, including spatial audio, motion information and long-duration videos. Secondly, the AI should be able to reason the space-time context in an underlying global coordinate space persistently. Thirdly, the AI should understand the wearer's intent by observing human eye gaze movement or physical interaction using hands. 

Empowering the research and applications in these areas will require datasets that are truly representative and contain sufficient contextual data to measure and train novel capabilities. Existing egocentric datasets are primarily captured using traditional video cameras and do not contain all the sensor modalities needed for this research. They usually lack the raw data expected from future AR glasses (such as spatial audio, inertial data), precise 3D location data, time synchronization between modalities and across devices, and the additional personal context to infer the wearer's action or intent. 

We introduce the Aria Everyday Activities (AEA) dataset, a 4D dataset with a rich suite of multi-modal sensory information and state-of-the-art machine perception outputs for AI and AR research. The AEA dataset was captured using Project Aria devices\cite{project_aria}, a sensor platform with unmatched capabilities for providing always-on egocentric data in an unobtrusive form factor. Aria data includes the following raw sensor data: high resolution RGB video, low resolution global shutter monochrome videos for location tracking, eyetracking videos, two IMU data streams, spacial audio from several microphones, magnetometer data and barometer data. We further provide outputs from our state-of-the-art Machine Perception Services (MPS) \footnote{\href{https://facebookresearch.github.io/projectaria_tools/docs/ARK/mps}{Machine Perception Services (MPS) documentation page}}. MPS data aims to provide highly accurate and reliable results that can be leveraged as contextual inputs to their models, or used as pseudo ground truth to measure objectives with more strict input constraints. 

AEA is a 4D longitudinal dataset with all daily activity recordings aligned spatial-temporally. AEA contains 143 activities recorded by multiple wearers in 5 locations that are representative for multi-person activities in a day. 
In additional to an on-device high frequency trajectory that provides 6DoF poses, AEA contains globally aligned 3D context for all observations in the same location. In each location, all the recordings from different wearers contain the aligned trajectories and a global point cloud in the same 3D coordinate system.
To indicate each wearer's intention, we provide the processed eye gazed calculated from the eyetracking camera streams. 
We further provide the time-aligned transcription of speeches in the recordings. For multi-person activities, all the recordings are time-synchronized accurately with external timecode, which ensures all the different observations are synchronized accurately not only in the space but also in the time domain.  

The AEA dataset is an updated version of the Everyday Activities section of the Aria Pilot dataset (APD) \cite{aria_pilot_dataset}, the first publicly released Aria dataset. This dataset has served as the first stop to attract researchers experimenting with Project Aria data in various research areas. Since APD's launch, Project Aria has been used in several new public datasets, including the Aria Digital Twin dataset\cite{adt}, the Aria Synthetic Environments dataset, and the Ego-Exo4D dataset\cite{egoexo4d}. Along these open data releases, Project Aria has improved the MPS and open-source tooling it provides\cite{project_aria}. 

In the AEA dataset, we provide three major updates to the original pilot dataset. First, we updated the data formats to be consistent with our new open-source tooling, allowing users to load and query data using a unified set of tooling across all Aria public datasets. Secondly, we reprocessed the raw Aria recordings with most recent Machine Perception Services, providing the following improvements: (1) more precise 6DoF pose information, (2) newly generated semi-dense point cloud with 2D observations, (3) more precise calibrated eyegaze with 3D directional vector, and (4) per-frame sensor extrinsic and intrinsic calibrations. Thirdly, we augmented our open-source tooling with easier data loading capabilities in C++ and Python, and created tools for running hands-on examples discussed in this paper. Among the existing Project Aria datasets, AEA will continue to serve as a unique dataset that enables researchers to study the egocentric everyday activities with all the recordings and machine perception context aligned in the same space and time coordinate frames.

\begin{figure*}[t]
    \centering
    \includegraphics[width=0.6\linewidth]{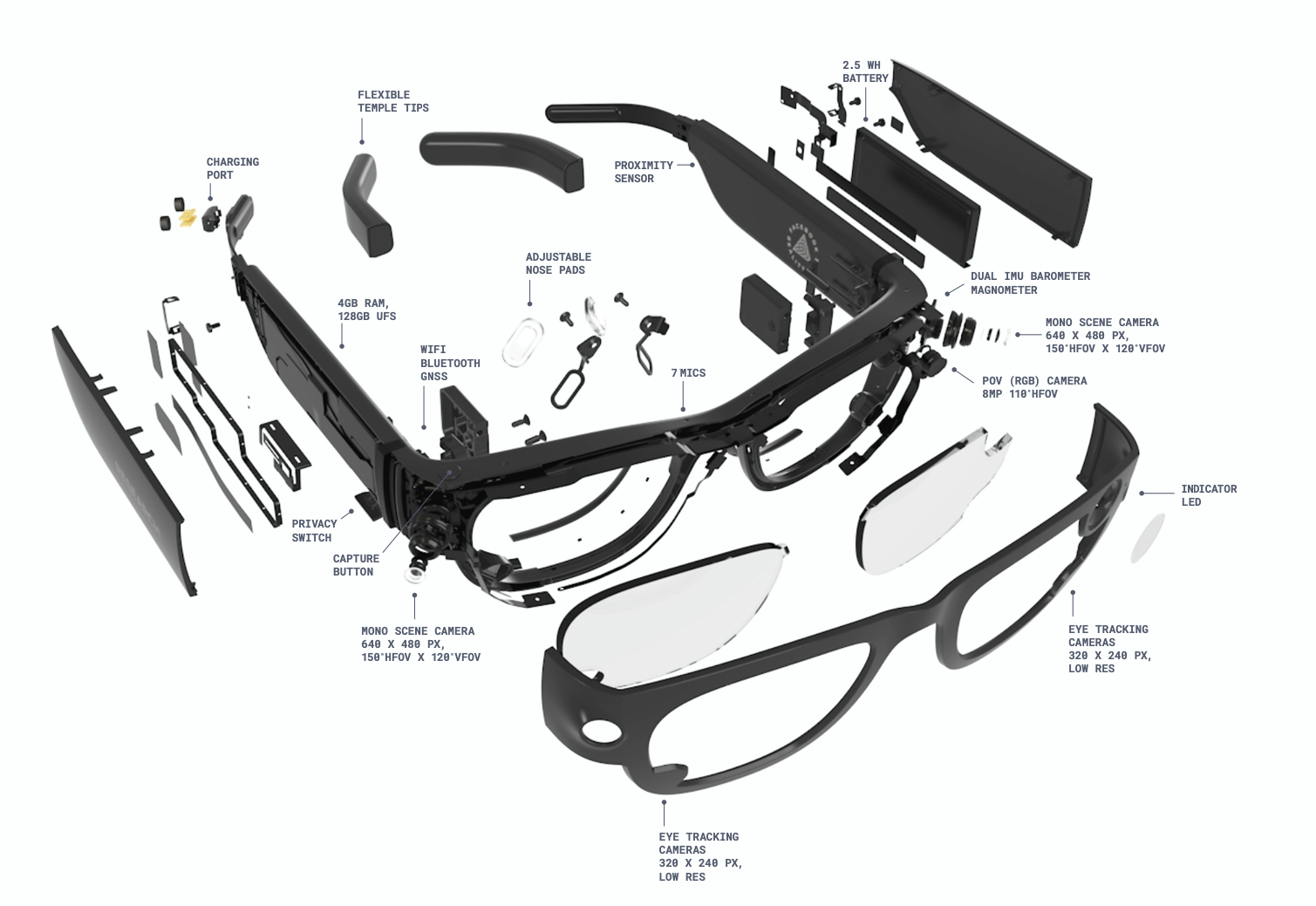}
    \includegraphics[width=0.35\linewidth]{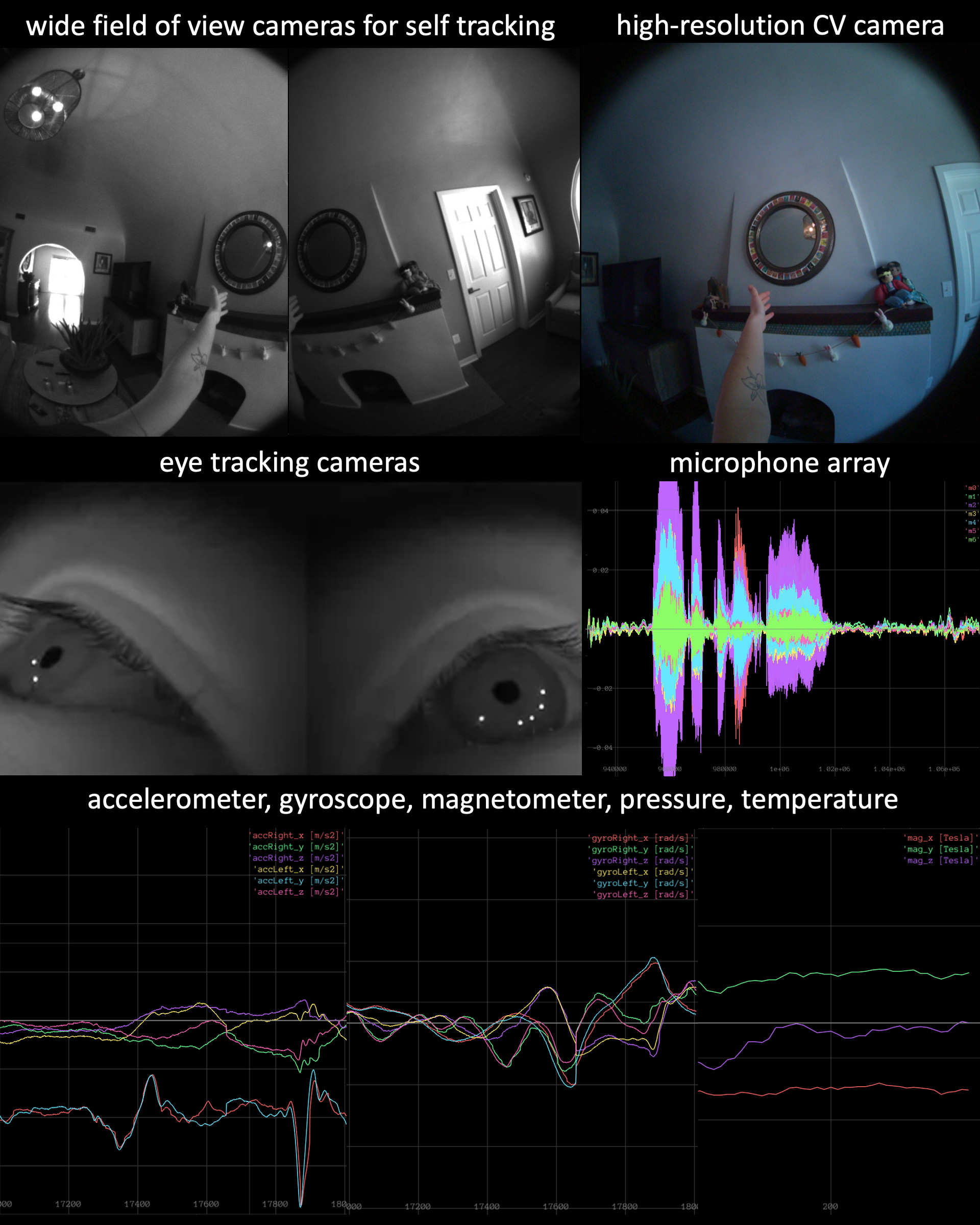}
    \caption{An overview of components on Project Aria device, with exemplar senor configuration of each. In this dataset, we use Profile9  which include all the sensors of the device except the WiFi, Bluetooth \& GNSS. We visualize a snapshot of all the sensors in one of the recordings on the right.}
    \label{fig:example_figure}
\end{figure*}

To further illustrate the capabilities of the AEA dataset in research, we have provided exemplar applications in neural reconstruction and prompted segmentation. In neural scene reconstruction, we demonstrate how to leverage the closed loop trajectories and global point cloud to reconstruct the observed scenes from either a single recording, or multi-recordings jointly using recent advances in Gaussian Splatting\cite{gaussian_splatting}. Finally, we've provided two prompted segmentation examples using Efficient-SAM\cite{efficientsam}, a distilled variant of the powerful foundational model SAM\cite{sam}, and grounding-DINO\cite{grounding_dino} to demonstrate how we can easily prompt the foundational model with wearer's input using eye gaze and speech. We hope these exemplar applications can inspire researchers to explore more powerful AI with this rich context. 

In summary, we present the AEA dataset, a multi-person everyday activities dataset recorded using Project Aria glasses, with multi-modal sensory data and 4D space-time aligned machine perception information. In this technical report, we will first discuss the related work in Sec.\ref{sec:related_work}. We will describe the major aspects of this dataset in Sec.\ref{sec:dataset_details} and provide open source tools in Sec.\ref{sec:dataset_tools}. Sec.\ref{sec:applications} provides a few example research applications using this dataset. 

\begin{table*}[t]
\centering
\begin{tabular}{@{}lcccccr@{}}
\toprule
Location         & Recording Scripts   & Wearers &  Recordings & Image Frames & RGB Frames & Total Duration (hours) \\ \midrule
Location 1 & Scripts 1-5       & 1-2 Wearers       & 29                   & 230,487          & 115,235              & 1.6                   \\
Location 2 & Scripts 1-5       & 1-2 Wearers       & 43                   & 339,650          & 169,824              & 2.3                   \\
Location 3 & Scripts 1-5       & 1-2 Wearers       & 38                   & 259,027          & 129,514              & 1.7                   \\
Location 4 & Scripts 1-5       & 1-2 Wearers       & 19                   & 94,029           & 47,015               & 0.6                   \\
Location 5 & Scripts 4-5       & 1 Wearer          & 14                   & 168,428          & 84,214               & 1.1                   \\ \midrule
In total & Scripts 1-5       & 1-2 Wearers          & 143               & 1,091,621          & 545,802              & 7.3     \\ \bottomrule
\end{tabular}
\caption{Data statistics for the recordings at each location. Refer to Appendix \ref{sec:activity_scripts} to more details of the recording scripts. The total number of image frames account for a combination of all RGB image frames and monochrome scene camera frames.}
\label{tab:APD_data_summary}
\end{table*}

\section{Related Work}
\label{sec:related_work}

In this section, we review prior work with a particular focus on datasets that have served the egocentric AI and 3D multimodal research community. 

\paragraph{Egocentric datasets:} The first-person vision (FPV) and egocentric datasets \cite{ego4d, egtea, epickitchens100, ego_procel} have served as an important role to drive research development in egocentric video understanding and AI. The two seminal datasets EpicKitchen\cite{epickitchens100, epickitchens_scaling} and Ego4D\cite{ego4d} set up a number of egocentric understanding benchmarks paired with extensive egocentric videos and annotations. These datasets were primarily captured by video recording devices mounted on wearer's head, e.g. a GoPro head-mounted camera rig. They lack other important sensor modalities required to more accurately infer 3D information or wearer intention. Due to the nature of rapid head-motion in egocentric videos, precisely and reliably estimating the 3D information in all the videos is still a very challenging research problem. A small subset of Ego4D videos rely on relocalization using a pre-scanned 3D environment. EpicFields\cite{epic_fields} recovered part of the 3D environments and trajectories using COLMAP\cite{schoenberger2016sfm} and neural fields. 
In the AEA dataset, we provide 6DoF 1KHz trajectories which approximate the continuous 6DoF trajectory of the egocentric observer for every single recording generated by the foundational capabilities of the Project Aria device using the state-of-the-art visual-inertial odometry (VIO) and simultaneous location and mapping (SLAM) systems. 
AEA also contains calibrated eye gaze information for every recording. Existing dataset GTEA\cite{gtea}, EGTea\cite{egtea}, and a small subset of Ego4D dataset contain eye-gaze information, but they do not contain other 3D information, which can create hurdles when contextualizing the wearers' intention in a 3D environment.  

\paragraph{Project Aria datasets:} AEA is an updated dataset based on the Everyday Activities sequences in the Aria Pilot dataset\cite{aria_pilot_dataset}. Since the Aria Pilot dataset's release, more datasets have been recorded using Project Aria devices that feature different capabilities and tasks. Aria Digital Twin (ADT) \cite{adt} contains real-world activities captured by Aria wearers in two environments that have been fully digitized with precise object and wearer poses. For all the Aria recordings, ADT contains the same raw sensor information, with additional ground truth rendered from the a motion capture system (MCS) aligned with the digitized scene. There are also some visual artifacts in the dataset to enable the MCS. Creating such a dataset requires extensive efforts and resources which can be difficult to scale up. 

In contrast, AEA was recorded in five diverse fully consented environments with no other requirements in the scene. We record multiple daily activities and acquired derived machine perception data using Project Aria's Machine Perception Services (MPS). AEA serves as example for Project Aria recordings that can be easily scaled up for research that requires extensive egocentric data capture. 

The recent effort of Ego-Exo4D dataset\cite{egoexo4d} features the largest multi-modal egocentric datasets using Project Aria as the egocentric recording device. It contains both egocentric and exocentric devices accumulated in 1400 hours. The dataset also uses MPS to generate per-recording machine perception data. The main differences between AEA and Ego-Exo4d are in the type of activities and environments being considered. Ego-Exo4d focuses on procedural activities, while AEA focuses on longitudinal daily activities in a whole day in home locations. In AEA, we provide multi-person synchronized egocentric recordings, and multi-trajectory recordings aligned in one home, while Ego-Exo4d provided synchronized captures between the egocentric videos with respect to the exocentric captures in a constrained cubic space observable from all cameras. We believe AEA can serve as an important pilot egocentric dataset for AI research that requires natural 3D environments and long-range spatial-temporal aligned videos. 

\paragraph{Other 3D multimodal datasets:} Recently, there have been a few other datasets introduced which were recorded using other head mounted devices in a more cumbersome form factor. Assembly101\cite{assembly101} uses Meta Oculus Quest devices and operate a series of assembly activities under a multi-view desktop environment. HoloAssist\cite{holo_assist} features several two-human collaborative tasks using the Hololens headset. Compared to these datasets, AEA features more natural human activities in indoor locations, with the benefits of non-obtrusive Project Aria form-factor. 

The study of multimodal research also extensively uses Autonomous driving datasets \cite{kitti, waymo, argoverse}, which provides very different scenarios and sensor modalities. The future of multimodal learning in personal AI will require datasets collected using similar device form-factors to future AR glasses and datasets that closely resemble the human activities in one's daily life. We believe AEA will serve as an exemplar dataset to drive research in this domain.

\section{Dataset Description}
\label{sec:dataset_details}

\begin{figure*}
    \centering
    \includegraphics[width=\linewidth]{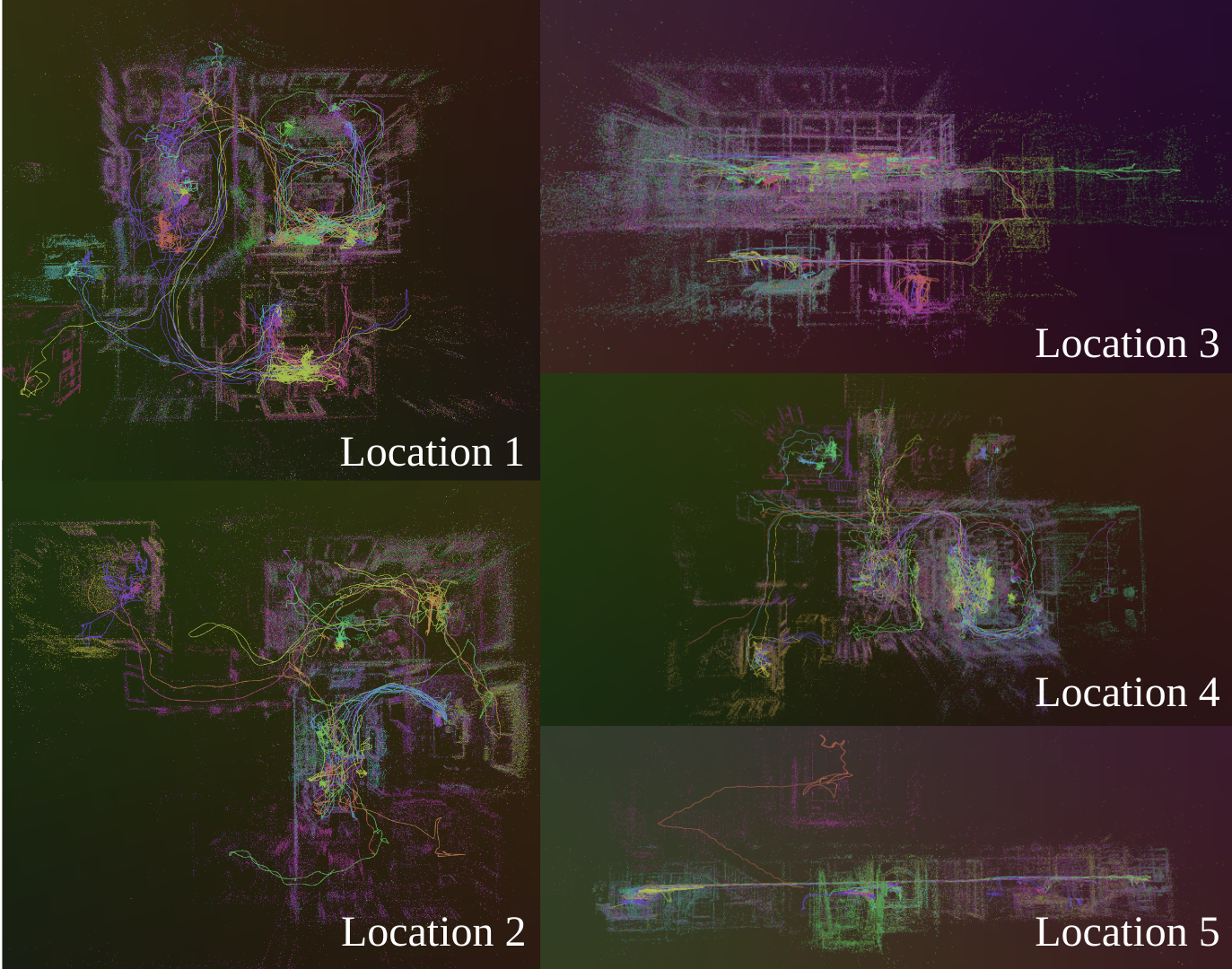}
    \caption{A visualization of the shared 3D global closed-loop trajectories and the semi-dense point clouds for multi-recording activities on every 5 location. Each color indicates the high frequency trajectory of one sequence recorded in this location. In each location, the point clouds are aggregated from all recordings. Location 3 and 5 are shown sideways to highlight the multi-floor scenarios.}
    \label{fig:multi_user_activity}
\end{figure*}

The Aria Everyday Activity (AEA) dataset is an updated version of Everyday Activity sequences within the Aria Pilot Dataset (APD). APD was the first publicly released dataset collected using Project Aria devices. It introduces a 4D longitudinal dataset with all day activity recordings captured by always-on form factor glasses with spatial-temporally aligned metadata. For the new AEA dataset, we updated all the machine perception data using the most recent Machine Perception Services (MPS) for all recordings, to provide the best 3D information possible. We describe the details of the AEA dataset below. 

\paragraph{Dataset statistics:} AEA contains 143 recordings of everyday activities collected by multiple wearers in five indoor locations. In total, it is composed of over 1 million frames with 7.3 accumulated recording hours. There are over 2 million image frames accounting for RGB color images and monochrome scene images. Tab.~\ref{tab:APD_data_summary} shows the data statistics for the number of wearers, recordings and frame numbers in each location. 

\paragraph{Dataset collection process:} We created guidance scripts for wearers to represent all day longitudinal activities that can be observed with always-on sensing for one to two Project Aria device wearers. Each script contained multiple scenarios that told a story about people going through their day. The scripts only provided general guidance for an improvised scenario. Wearers followed the provided prompts and went with what felt most natural to them in each of the recordings. The scripts can be used as an open-vocabulary description for the egocentric activities. Some of the scripts scenarios are: cooking, daily cleaning, exercising, dining, reading, playing games and spending time with friends. 
We provide details about how the scripts correspond to specific tasks in Appendix \ref{sec:activity_scripts}. 

\paragraph{Aria raw sensor data:} We used the \emph{recording Profile 9} provided by Project Aria data collection configurations, with 20 fps RGB camera at $1408\times1408$ resolution, 10 fps monochrome scene cameras at $640\times480$ resolution, 10 fps eye tracking camera at $320\times240$ resolution, and all other sensors configured at their default settings. Refer to Appendix \ref{sec:raw_sensor} for more details of the raw senor description.  

\paragraph{Machine perception data:} This is a major feature update in AEA dataset compared to recordings in Aria Pilot Dataset. AEA uses the current Machine Perception Services (MPS) provided to \href{https://www.projectaria.com/research-kit/}{Project Aria academic partners}. The data format has been updated to be consistent with the new official \href{https://facebookresearch.github.io/projectaria_tools/docs/data_formats}{Project Aria data format convention} used in Project Aria Tools, the new opens source tooling and documentation repository. There are two major updates to the MPS data compared to the Aria Pilot dataset,
\begin{itemize}
\itemsep0em 
    \item \textbf{3D location}: We've provided 6DoF trajectories aligned with visible global point cloud for each recording, which shares the same coordinate system with all other recordings in the same location. The global point cloud is a reconstruction using tracked feature points of the static portion of the environment. We call the 3D point cloud as \textit{semi-dense points} and the 2D tracked points as \textit{semi-dense observations} in MPS. We only provided 1Khz trajectory for each recording in the initial Aria Pilot Dataset. In this update, we have added semi-dense points aligned with each trajectory in the global coordinate space, which we call \textit{close-loop trajectory}. Fig.~\ref{fig:multi_user_activity} shows the visualization of all spatial aligned recordings in each location. We've manually inspected the trajectories and point cloud for all recordings in the global aligned coordinate to ensure they are correct. 
    \item \textbf{Eye gaze with 3D directional vector }: AEA provides calculated 3D eye gaze as a per-frame 3D ray anchored on the Central Pupil Frame. This provides more information in the 3D environment compared to the initial release of 2D reprojected eyegaze point in RGB camera. We've also provided code examples in Project Aria Tools to acquire the 2D eye gaze using device calibration.

\end{itemize}
We've also provided on-device odometry, which we call \textit{open-loop trajectory} and online device calibration. Both of these features were not available when we released the pilot datasets. For more details about the machine perception data and MPS, please refer to Appendix \ref{sec:mps}.

\paragraph{Time synchronization:} AEA includes several multiple-person activities, such as two people having a conversation. In multi-person scenarios, we captured the recordings with synchronized timestamps. Multiple Project Aria devices that operate in proximity to each other ($<100m$) can leverage SMPTE LTC timecode to receive a synchronized time clock with sub-millisecond accuracy. In the VRS file, we've provided synchronized timecode timestamps shared across multi-device captures in addition to the capture time of each device. Different recordings captured at the same time can be associated using the device timecode. We also provide a toolkit to visualize the synchronized recordings, which we will illustrate in Sec.~\ref{sec:dataset_tools}. 

\paragraph{Speech2text transcription:} We've used the same text transcription generated by Automatic Speech Recognition (ASR) released in APD. All characters in text are aligned with the device timestamp, and each output is also associated with a confidence rating. The ASR annotation used a proprietary service that is not part of Project Aria MPS. Similar results can be acquired via open-source ASR solutions, such as \cite{seamless2023, whisper}. We hope the released speech transcription can accelerate the research and applications that require time aligned text input. 

\begin{figure}
    \centering
    \includegraphics[width=\linewidth]{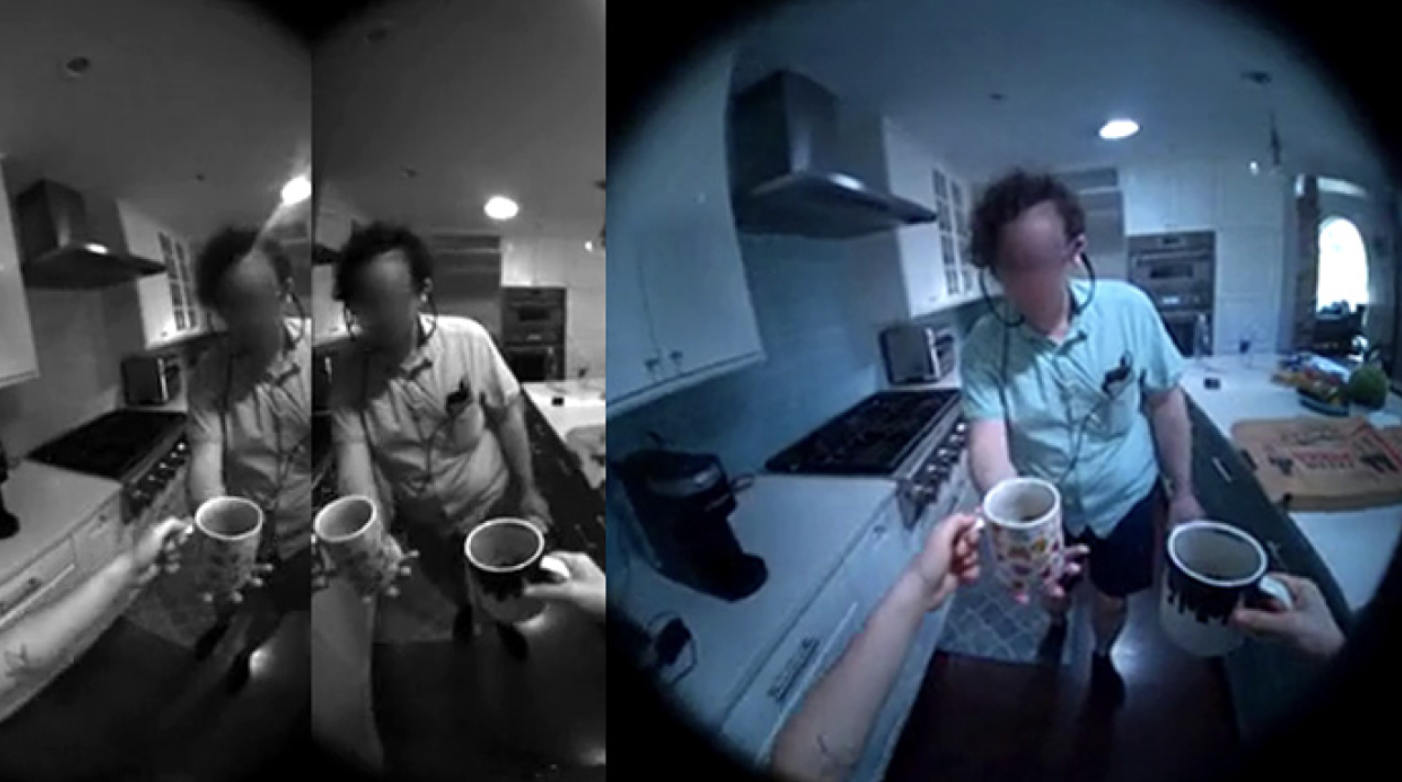}
    \caption{We manually blurred all the human faces in both RGB color video (right) and two monochrome scene camera videos (cropped in the left two images).}
    \label{fig:anonymization}
\end{figure}

\paragraph{Privacy commitments:} We strictly follow the Meta's Responsible Innovation Principles for all data collection and processing. All of the recordings were captured in fully consented indoor environments with no personal identified information. To ensure the highest quality of anonymization, we manually annotated and blurred all the faces and licenses on all images in RGB and monochrome scene camera streams. We carefully verified the annotation results with multiple rounds of quality analysis. Recordings flagged with frames that did not pass quality check were completely excluded from the dataset. Fig~\ref{fig:anonymization} shows an example of the anonymization for RGB and monochrome sensor streams.  

\section{Dataset Tools}
\label{sec:dataset_tools}

\begin{figure*}[t]
    \centering
    \includegraphics[width=\linewidth]{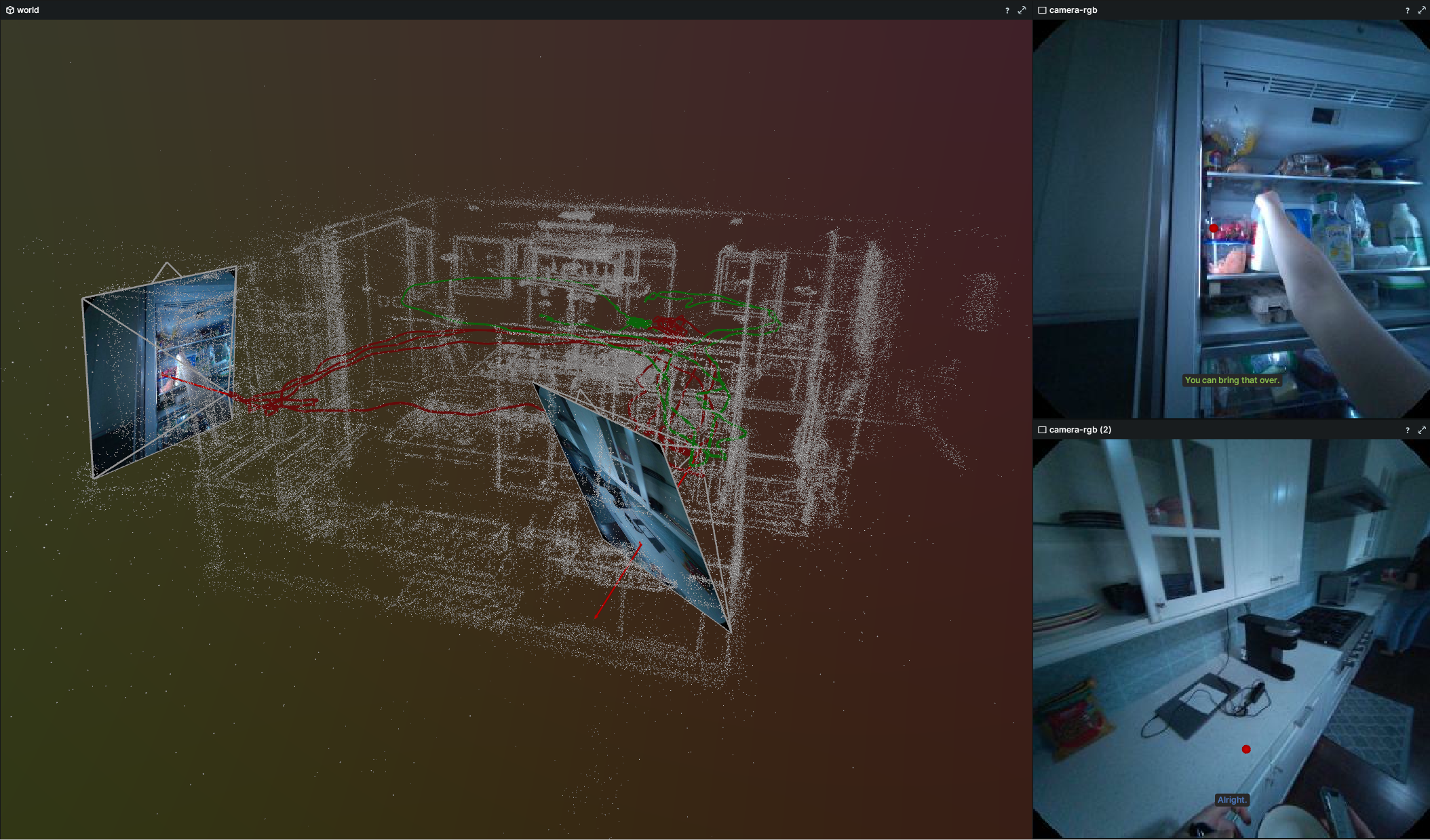}
    \caption{A snapshot of AEA dataset viewer (multiple-person activity) showing the synchronized rectified RGB streams with their devices trajectories (dark red and green of each), eye gaze vector direction (a red vector from each device frustum), the projected eye gaze on RGB image (red dot in each image), the aggregated 3D semi-dense point cloud for each recording (white) and the transcribed speech sentences (overlaid on each RGB image). We provide this viewer in project aria tools.}
    \label{fig:tool_viewer}
\end{figure*}

To support the release of this dataset we updated Project Aria Tools, which provides the open source data utilities for working with Aria data. This update provides data providers to consume the multimodal sensory data and their paired machine perception data from multiple devices. All data modalities and metadata can be retrieved by their device \textit{timestamp}. Manipulating single or multiple-person activities can be done with ease. Project Aria Tools also provides convenient APIs to use the device \& camera calibration to manipulate the 6DoF pose transformation as well as 2D/3D points projection and reprojection between the various sensors.

Project Aria Tools provides a visualization tool to support users quickly exploring each sequence or multi-person scenario with temporal scrubbing. Fig.~\ref{fig:tool_viewer} shows an example of two time synchronized recordings with their 3D trajectory, environment as global point cloud and eye gaze vectors visualization all in one shared coordinate. We also display the synchronized text from speech as an example of other temporal data, which is queried according to the device time. This could be easily extended to display additional user metadata or more recordings for development and debugging.

\section{Applications}
\label{sec:applications}

We provide two main research applications of the AEA dataset in this section, 3D neural scene reconstruction, and prompted segmentation. These exemplar applications demonstrate strong use cases for using AEA multimodal egocentric observation and machine perception data. We hope they can give a glimpse of research using this dataset. 

\subsection{3D Neural Scene Reconstruction}

Reconstructing scenes for observable areas is one important problem to create immersive photo-realistic virtual memories in AR/VR applications. Existing neural reconstruction methods usually require a two-step approach. Given one video sequence, they acquire the 6DoF camera trajectory from a structure-from-motion (SfM) tool, e.g. COLMAP \cite{schoenberger2016sfm}, and then run neural reconstructions on the posed images. This can be challenging for egocentric activities, where the recordings are not carefully taken and curated for the purpose of running SfM. Rapid head motion or less observable areas can easily lead to failure of SfM or similar localization algorithms, and fail further reconstructions. For multiple long recordings in the same environment, it can also be challenging to jointly localize all the observations in a shared coordinate system. 

It is worth noting that the AEA dataset is activity-centric. Every recording contains various egocentric motions and may also observe moving humans in multi-person activities. All of the trajectories are free-view motions that are drastically different from traditional datasets made for object or static scene oriented reconstruction purposes. Despite these challenges, we demonstrate that we can successfully reconstruct scenes that are persistent across one or multiple recordings. We hope this can inspire future research to improve the reconstruction quality when there are dynamic motions, or improve activity \& scene understanding with fully immersive environments.

In this section, we demonstrate running neural reconstruction and address the aforementioned challenges using 3D machine perception data from the AEA dataset, in particular the closed-loop trajectory and semi-dense point cloud. We use Gaussian-splatting \cite{gaussian_splatting} as the primary baseline and provide a full evaluation. 

\begin{figure*}[h]
    \centering
    \begin{tabular}{c}
    rendering views \\
    \includegraphics[width=0.24\textwidth]{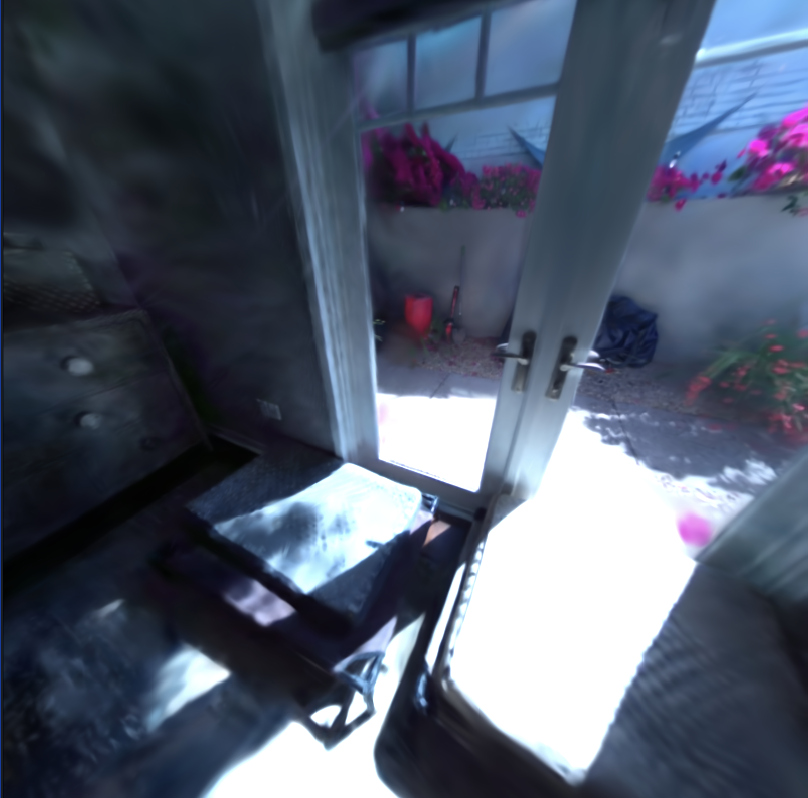}
    \includegraphics[width=0.24\textwidth]{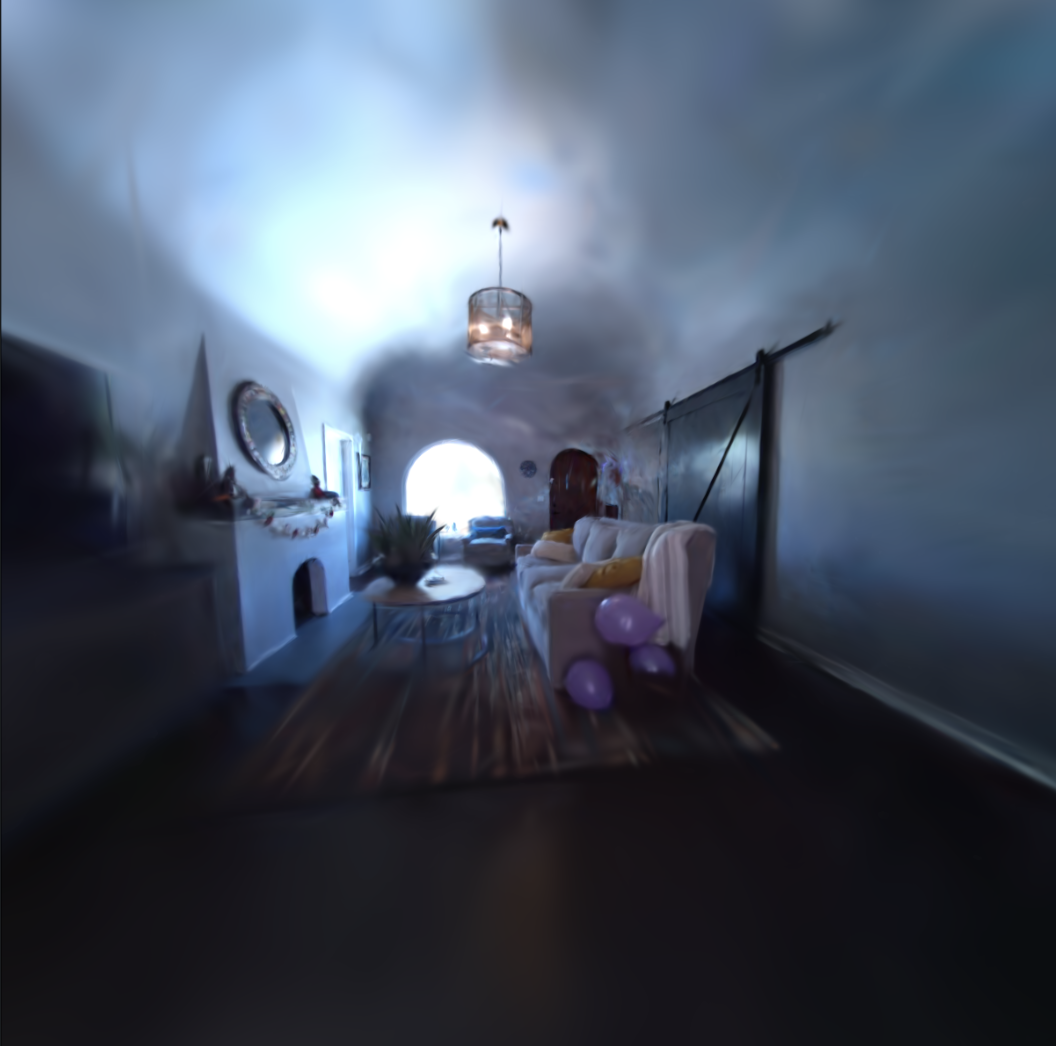}
    \includegraphics[width=0.24\textwidth]{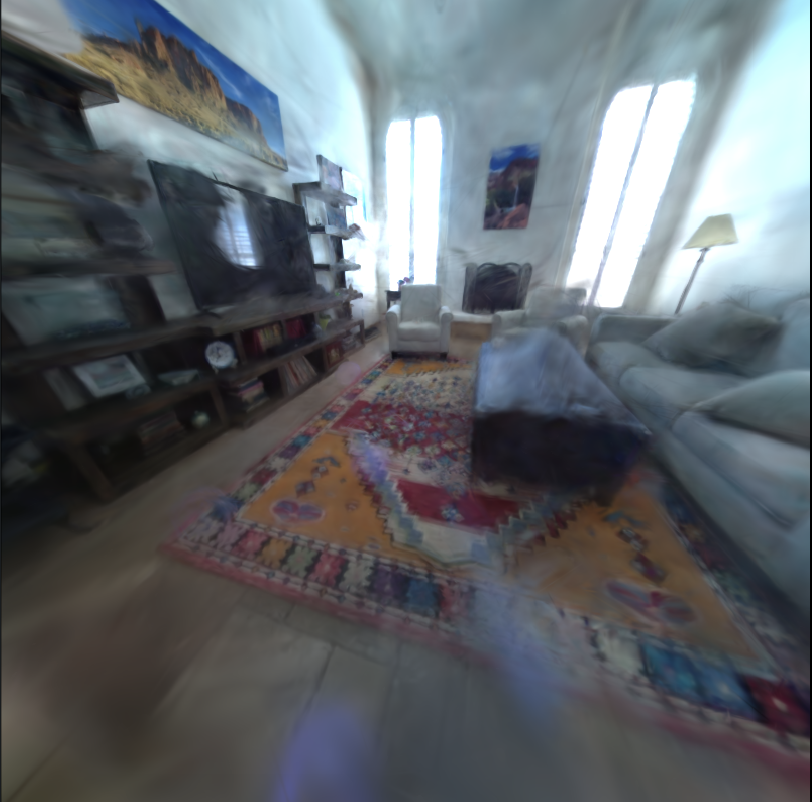}
    \includegraphics[width=0.24\textwidth]{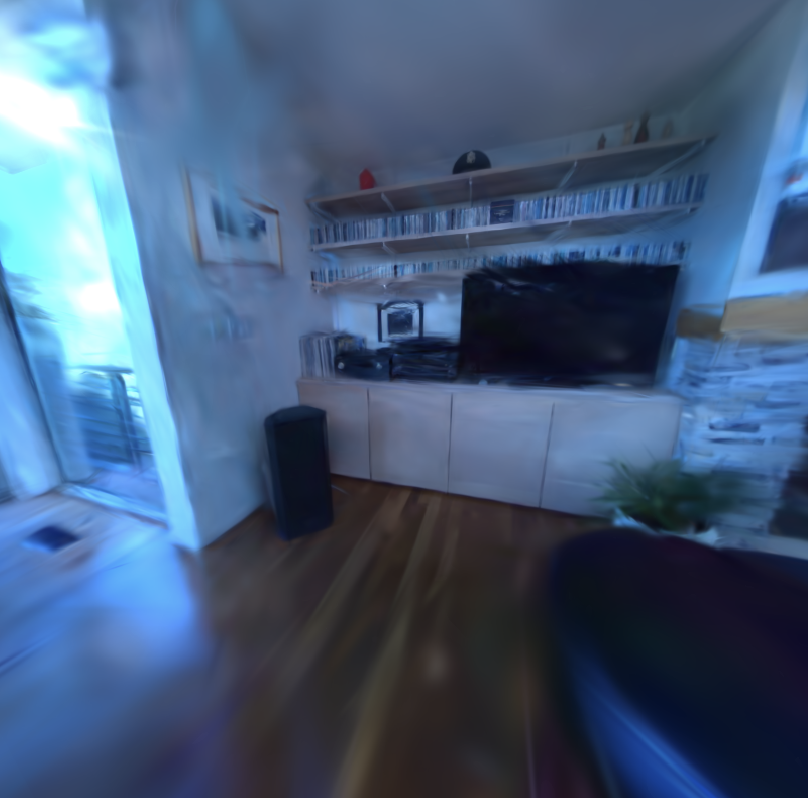} \\
    \hline
    nearest neighbour views corresponding to above rendering views \\
    \includegraphics[width=0.24\textwidth]{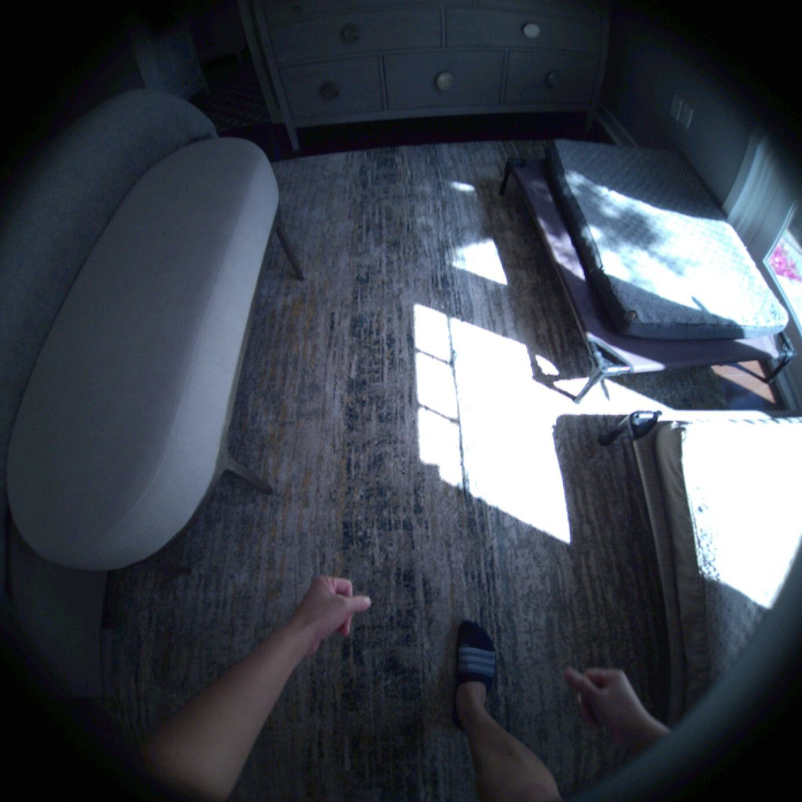}
    \includegraphics[width=0.24\textwidth]{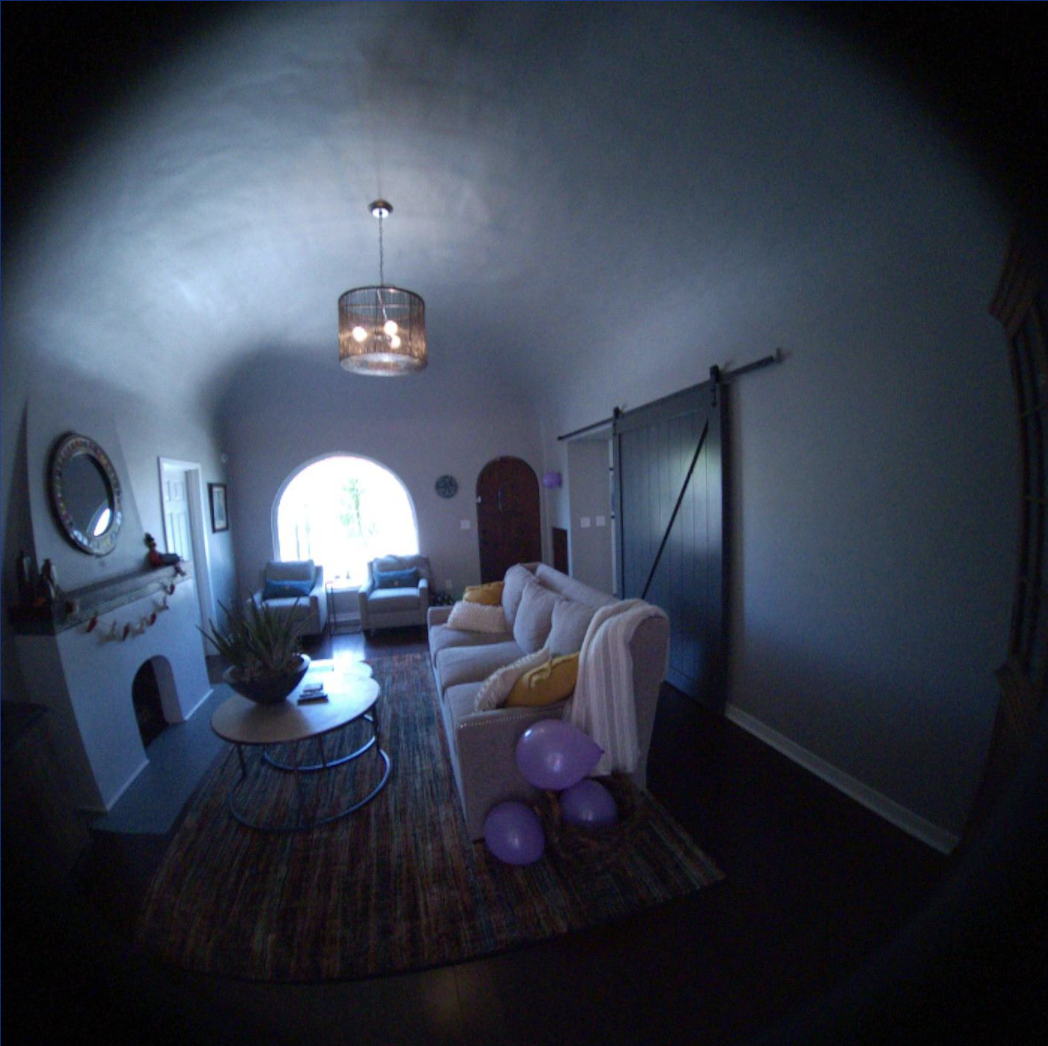}
    \includegraphics[width=0.24\textwidth]{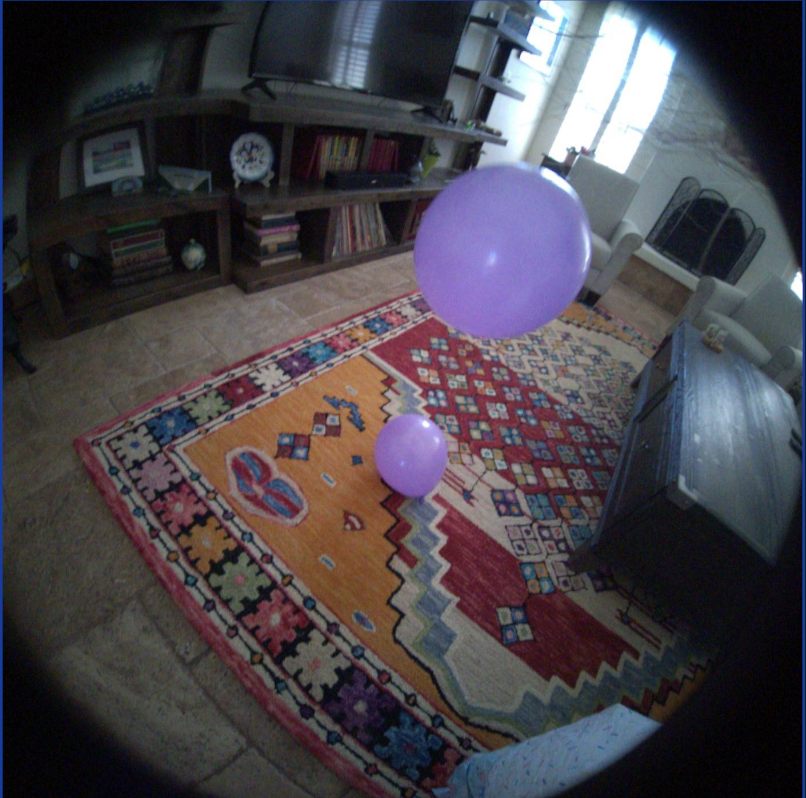}
    \includegraphics[width=0.24\textwidth]{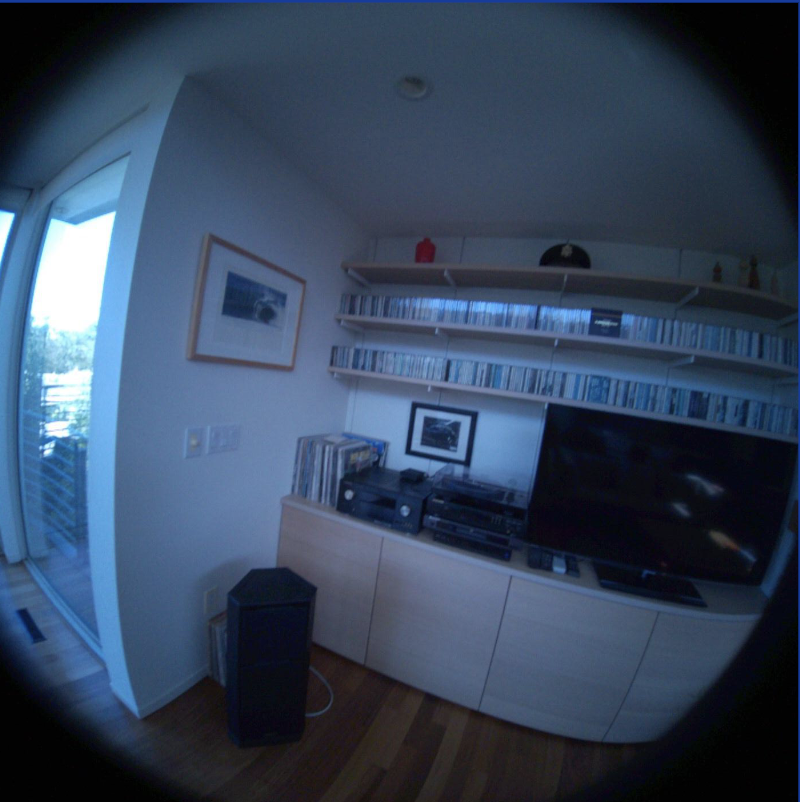}
    \end{tabular}
    \caption{Novel rendering views of the Gaussian-splatting reconstruction (top) and their corresponding nearest neighbor views (bottom) using observations \textbf{from a single recording}. }
    \label{fig:gs_single_recording}
\end{figure*}
\begin{figure*}[h]
    \centering
    \begin{tabular}{c}
    rendering views \\
    \includegraphics[width=0.24\textwidth]{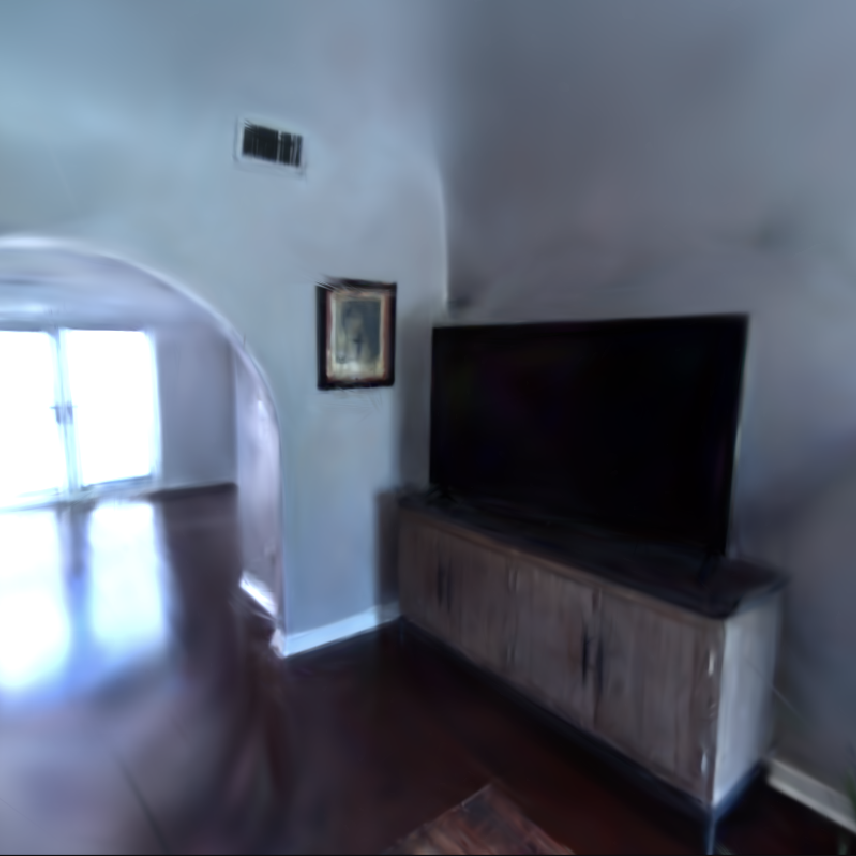}
    \includegraphics[width=0.24\textwidth]{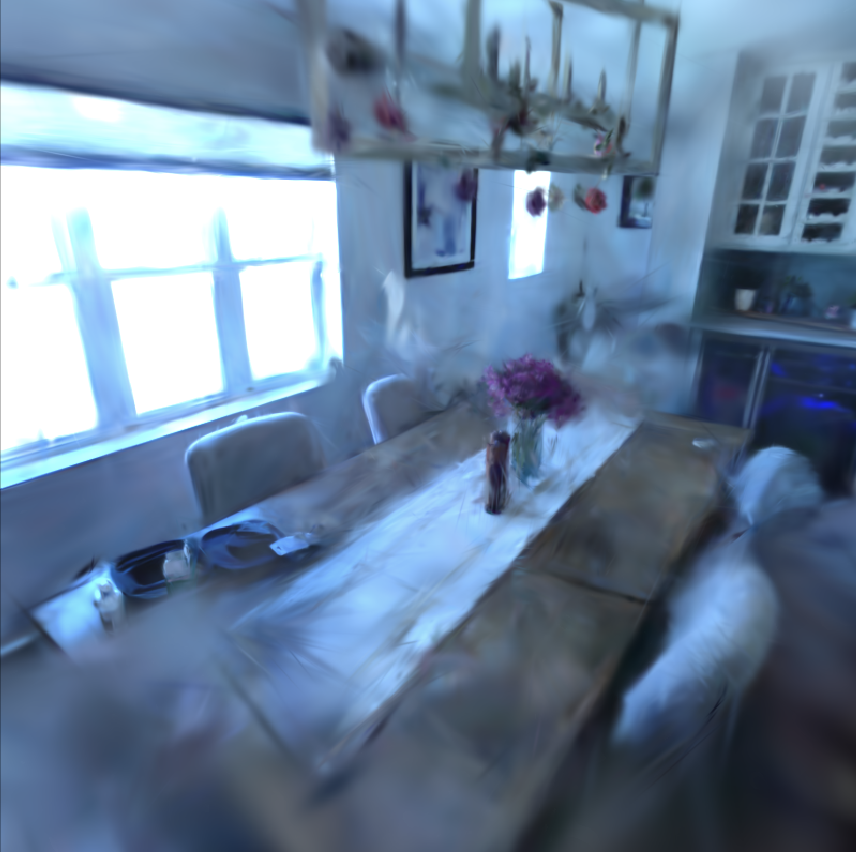}
    \includegraphics[width=0.24\textwidth]{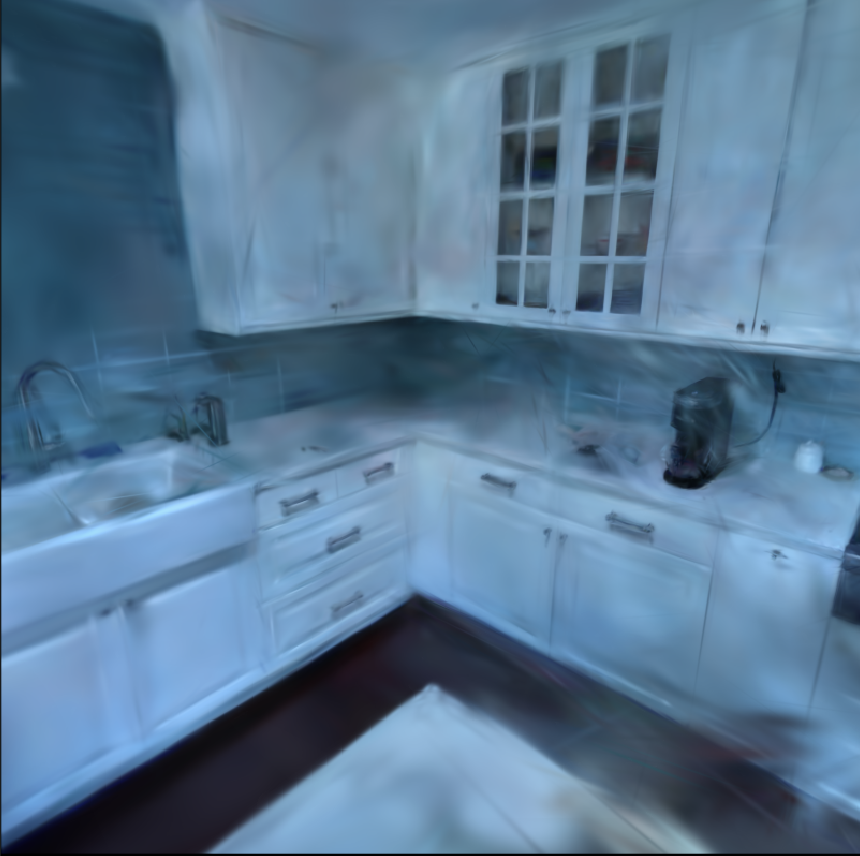}
    \includegraphics[width=0.24\textwidth]{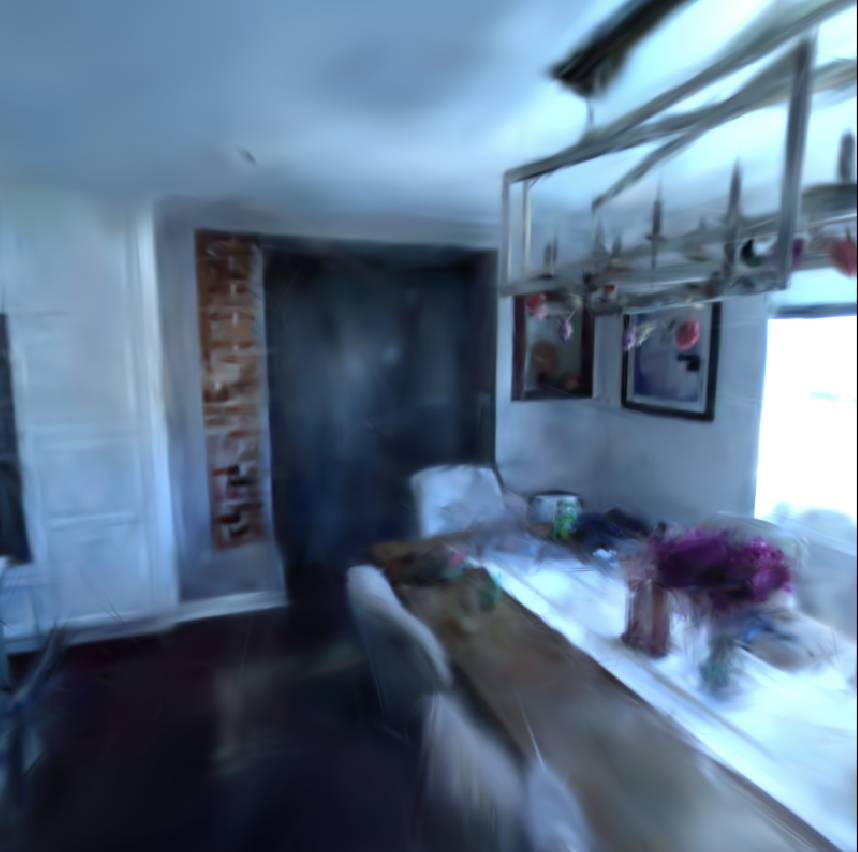} \\
    \hline
    nearest neighbour views corresponding to above rendering views \\
    \includegraphics[width=0.24\textwidth]{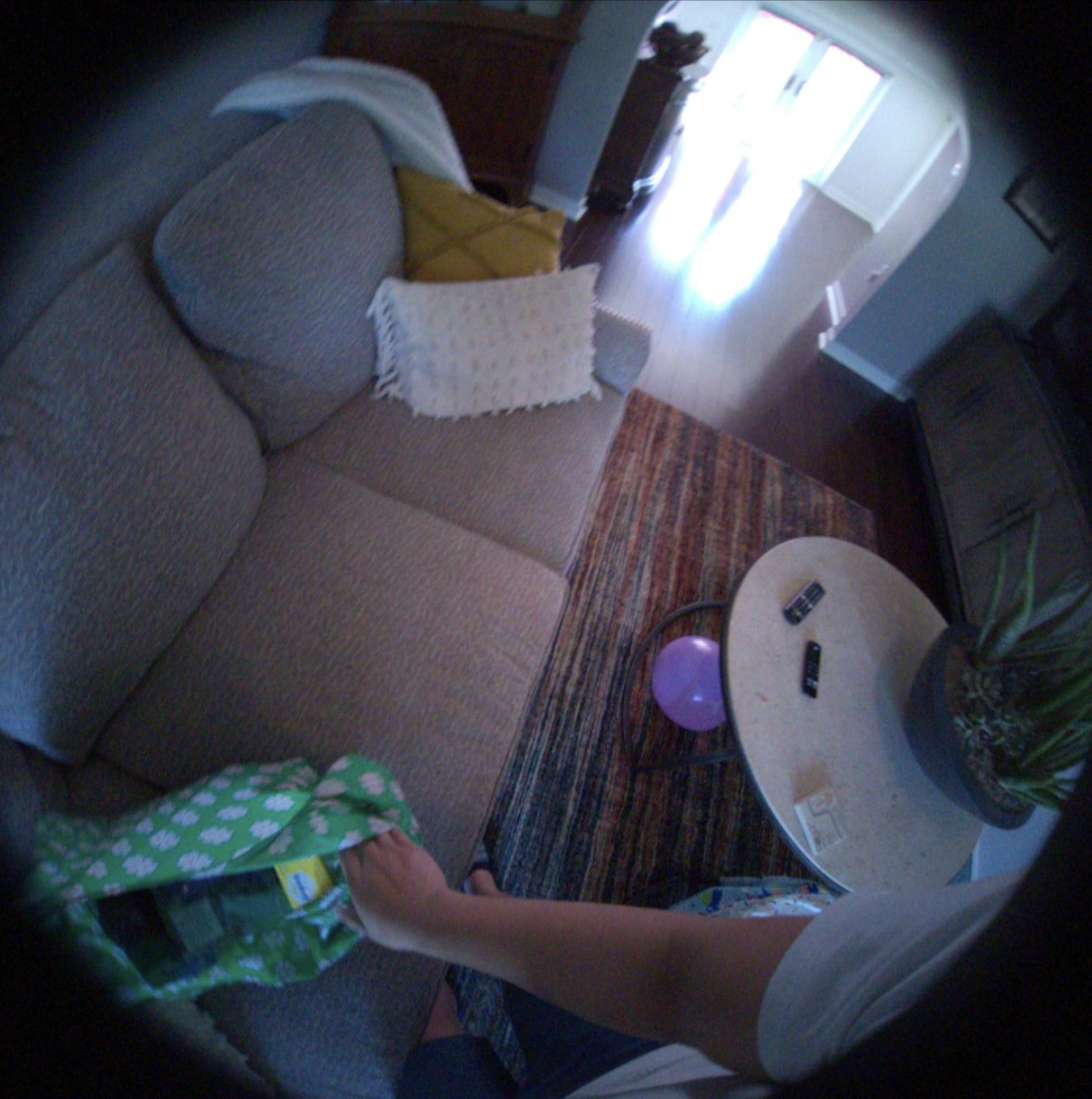}
    \includegraphics[width=0.24\textwidth]{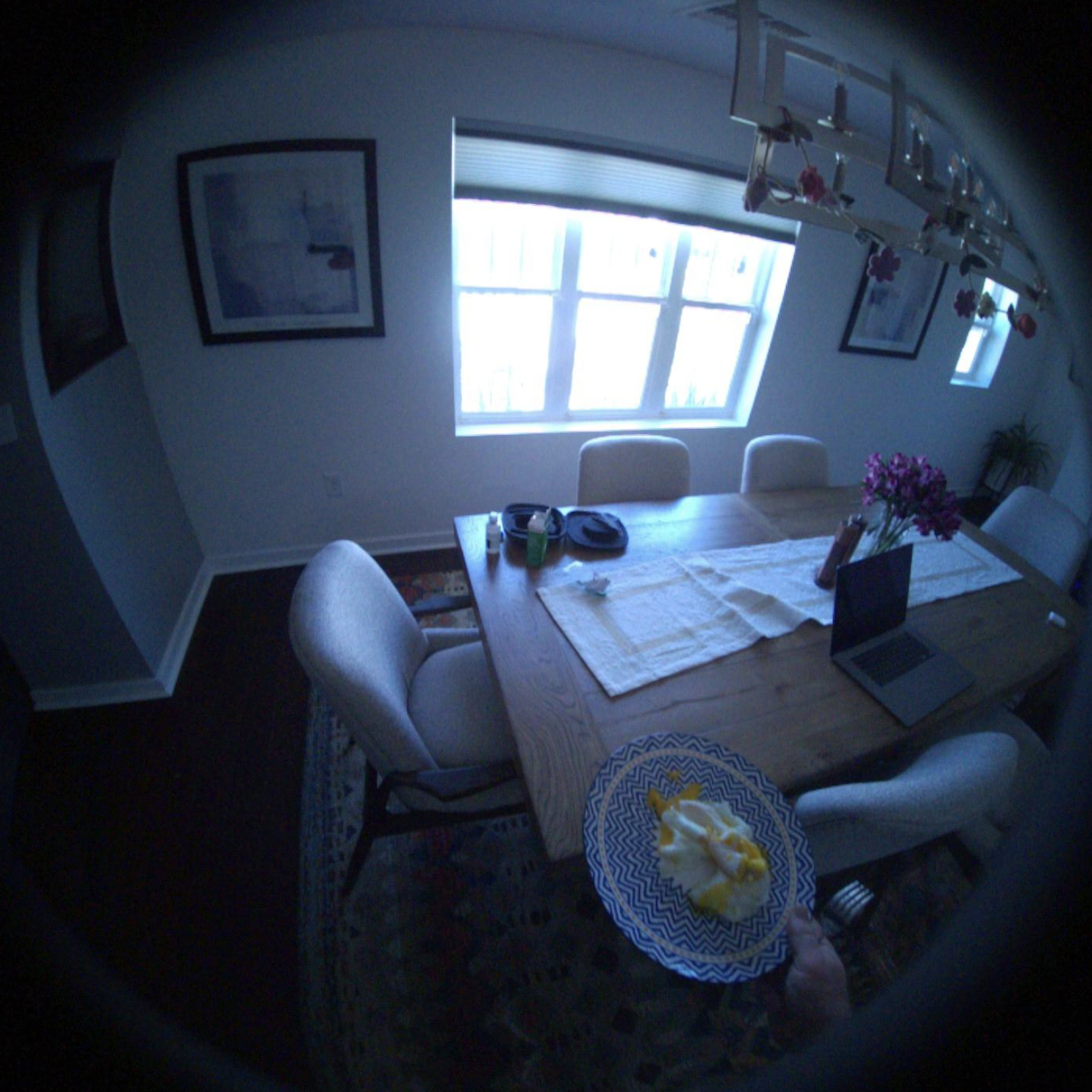}
    \includegraphics[width=0.24\textwidth]{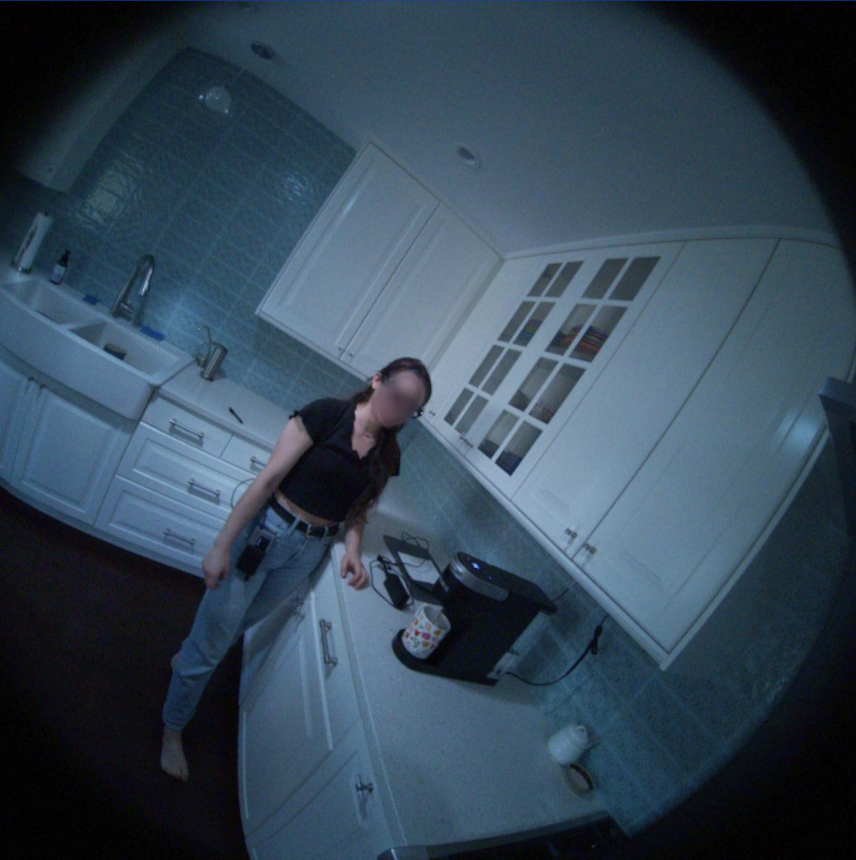}
    \includegraphics[width=0.24\textwidth]{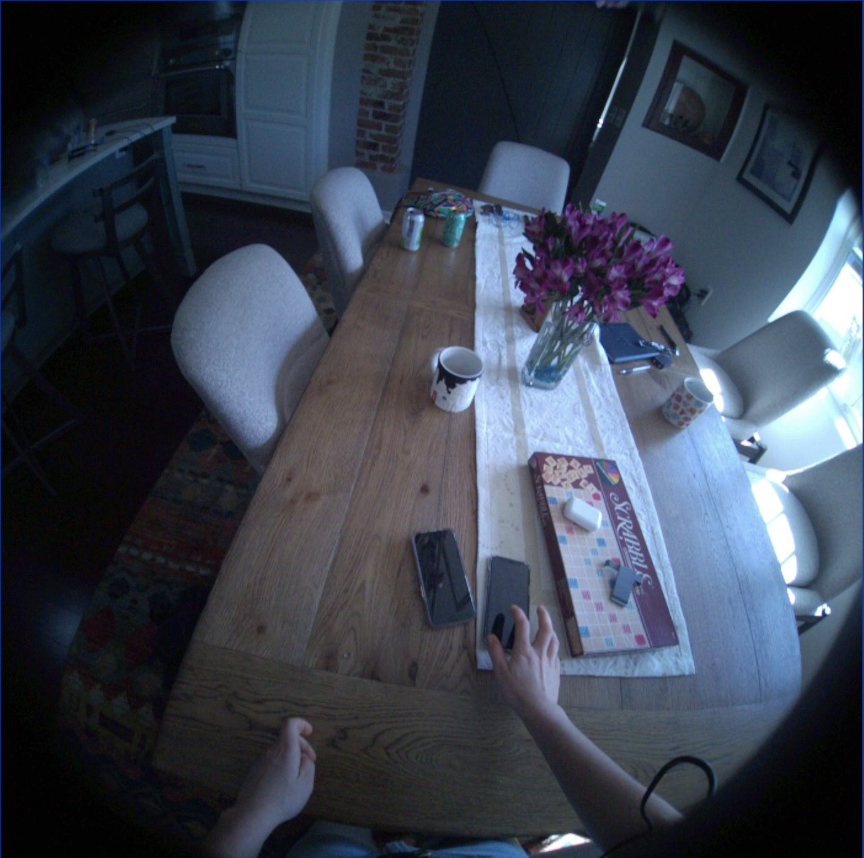}
    \end{tabular}
    \caption{Novel rendering views of the Gaussian-splatting reconstruction using observations \textbf{from all recordings in the shared location}. The corresponding nearest neighbour views are shown on the second row.}
    \label{fig:gs_multiple_recordings}
\end{figure*}

\subsubsection{Gaussian splatting}

\paragraph{Implementation:} We followed the official implementation of Gaussian Splatting, except for a few minor changes to support multi-sequence reconstruction. Since we reconstructed long trajectories which could not be loaded directly into memory, we used batch processing to randomly seek a batch of frames at runtime. We used Project Aria Tools to rectify all the RGB images that fit in the requirements of the default pinhole rasterizer. We initialized the point cloud using the AEA semi-dense points, and acquired per-frame poses using the pose with the nearest timestamp from the high-frequency closed-loop trajectory. For multi-recording reconstructions, we concatenate all the frames, and point clouds together. We ran 30K iterations for single-recording reconstruction and 100K iterations for multi-sequence reconstructions. All reconstructions shared the same hyper-parameters and strategy for pruning and densification. On a single A100 80GB GPU, reconstructing a single recording took about 3 hours, and 10 hours for multiple recordings in one shared location.

\paragraph{Single Recording Results:} This method successfully converged on each recording individually, despite some recordings having high dynamic motion or minimal head motions. Fig.~\ref{fig:gs_single_recording} shows a few rendering examples with the nearest view in the 3D space. In the results, we saw good quality reconstruction for all well-observed static areas. The blurry areas rendered from Gaussians indicate the dynamic or unobserved areas in the original observation. 

\paragraph{Multi-recording Results:} We further ran Gaussian Splatting on all recordings aggregated in each location. The number of recordings and frames can be found in Tab.\ref{tab:APD_data_summary}. For example, location 2 contained the longest recordings accumulated with total length of 2.3h, close to 170K RGB frames, and 5M initial points aggregated from the semi-dense point clouds. After 100K iteration of reconstruction, location 2 converged to about 627K Gaussian points. Fig.~\ref{fig:gs_multiple_recordings} illustrates a few rendered views. Given the abundance of observations in one location, we could reconstruct well-observed living areas and remove the image areas with dynamic motions as outliers in each. 

\paragraph{Quantitative Evaluations:}  Tab.~\ref{tab:gs_evaluation} shows the quantitative reconstruction quality evaluated for each location, in single recording (SR) and multi-recording (MR) settings. We reported the PSNR number averaged for all the held-out test frames. The test-frames were the last-frame out of every 5 consecutive frames. 

\begin{table}[t]
\centering
\begin{tabular}{@{}llccccr@{}}
\toprule
Location & PSNR (SR) & PSNR (MR)  \\ \midrule
Location 1 & 25.79 & 22.85  \\
Location 2 & 25.19 & 21.93  \\
Location 3 & 24.85 & 22.06  \\
Location 4 & 23.70 & 21.83  \\
Location 5 & 26.27 & 24.18  \\ 
Locations all & 25.12 & 22.35 \\ \bottomrule
\end{tabular}
\caption{Quantitative evaluation of Gaussian-splatting on the held-out testing frames (The last frame out of every 5 frames). We report the PSNR averaged for all frames in single recording (SR) or multiple recordings (MR) scenarios.}
\label{tab:gs_evaluation}
\end{table}

\subsubsection{3D Reconstruction using NeRFstudio}

NeRFstudio~\cite{nerfstudio} provides an \href{https://docs.nerf.studio/quickstart/custom_dataset.html#aria}{Aria customized integration} for processing Aria raw data using the trajectories and global point cloud provided by MPS. 
The NeRF baselines, e.g. NeRFacto use the Aria closed-loop trajectories and raw RGB images as input. Its Gaussian-splatting baseline (aka Splatfacto) also uses semi-dense point cloud data to initialize the 3D Gaussians and rectify the RGB images as we demonstrated. To achieve the best results on AEA, some adjustments in the current implementations were needed to accommodate the length of recordings. We will continue to support open-source implementation of scene reconstruction using Project Aria data, and plan to reconstruct AEA dataset with baselines in NeRFstudio in the future work.

\subsection{Prompted Segmentation}

The ability to recognize and segment object instance in the 3D environment is one of the fundamental components in contextual AI applications. The recent progress in 2D foundational models show a promising path to enable zero-shot AI capabilities with good generalization, e.g. CLIP\cite{clip}, DINO\cite{dino}, SAM\cite{sam}. In this section, we use segmenting anything model (SAM) as an example to show multiple sensor modalities can be helpful for AI research. We use the EfficientSAM\cite{efficientsam}, which is a variant of SAM, as it provides a good balance between accuracy and real-time performance when running on every RGB frame. 

In this example we show two applications connecting EfficientSAM with two different prompts from the machine perception data. First, we show how to segment objects using the eye gaze as the prompt of wearer's intention. Then, we further connect EfficientSAM with GroundingDino\cite{grounding_dino} to demonstrate speech grounded segmentation. Both applications are provided as implementation examples in the Project Aria Tools.

\begin{figure}
    \centering
    \begin{subfigure}[b]{0.48\linewidth}
        \centering
        \includegraphics[width=\linewidth]{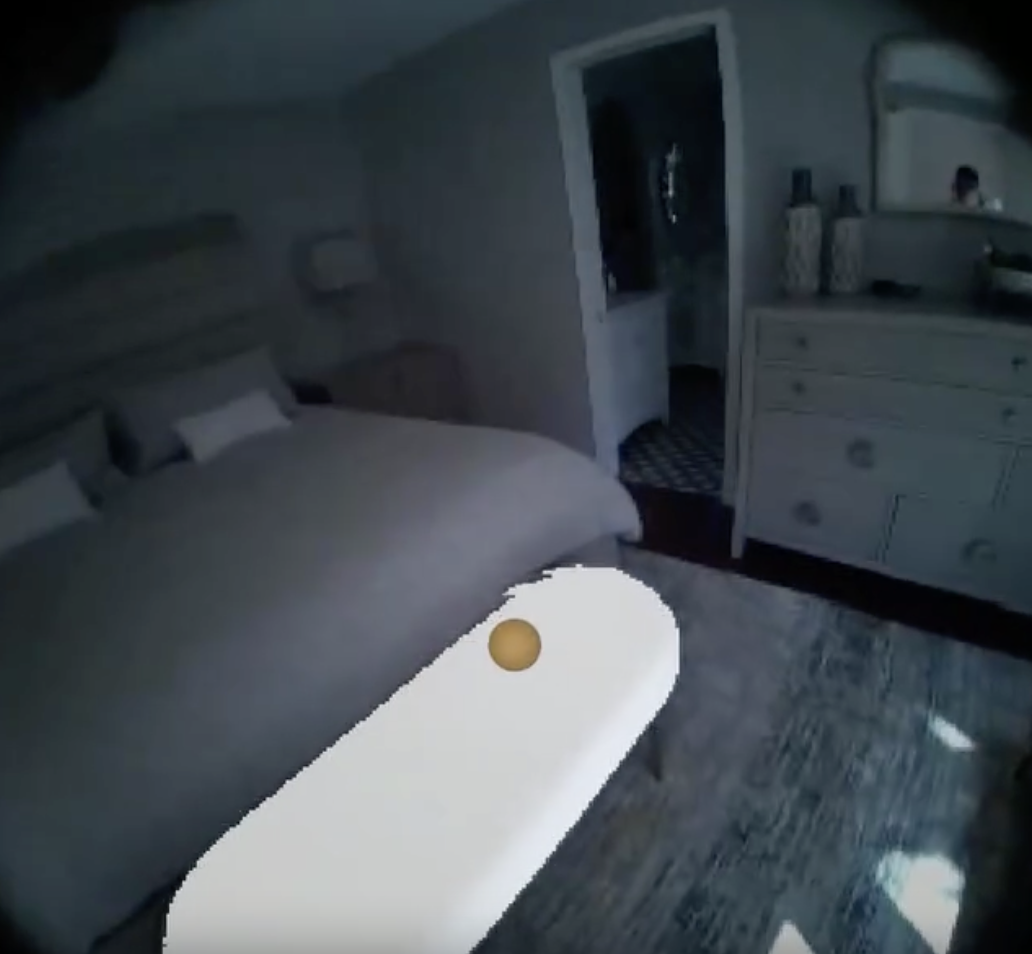}
        \caption{Eye gaze prompt}
        \label{fig:eye_gaze_prompt}
    \end{subfigure}
    \begin{subfigure}[b]{0.45\linewidth}
        \centering
        \includegraphics[width=\linewidth]{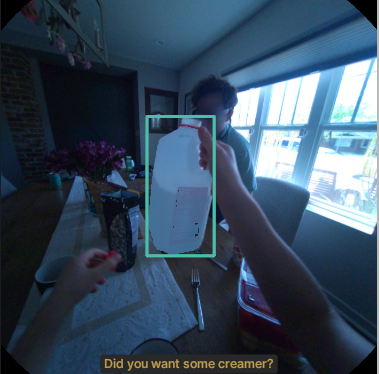}
        \caption{Speech prompt}
        \label{fig:speech_prompt}
    \end{subfigure}
    \caption{We demonstrate multimodal prompted segmentation to highlight what the wearer is looking at or talking about. The eye gaze projection is used as point prompt to EfficientSAM \cite{efficientsam} in (a) and speech is used to prompt GroundingDino\cite{grounding_dino} to detect the object and further segment them out using EfficientSAM in (b).}
    \label{fig:prompted_segmentation}
\end{figure}

\paragraph{Eye-gazed prompted segmentation:} Fig.\ref{fig:eye_gaze_prompt} shows the visualization of segmentation using projected eye-gaze in one RGB image as a prompt. Given the eye gaze directional vector and a virtual depth along this vector, we projected it into 2D RGB image stream using the device calibration, and then used the 2D reprojected eye gaze as the location prompt to produce a segmentation mask with EfficientSAM. In this example, we set the virtual depth at 1.0 meter. Future work can improve prompt accuracy by incorporating a more accurate eye gaze depth estimator.

\paragraph{Speech grounded segmentation:} Fig.\ref{fig:speech_prompt} shows a visual grounding of an object mentioned in the speech. Given a time-aligned speech sentence, we first used GroundingDino\cite{grounding_dino} align the text with the image and detect the objects in the sentence. If an object existed, the corresponding bounding box was used as the box prompt and EfficientSAM\cite{efficientsam} produced a segmentation mask within it.

\section{Conclusion}
\label{sec:conclusion}

We introduce the Aria Everyday Activity (AEA) Dataset for the research community to explore multimodal AI research with real world personal longitudinal activities that have spatial-temporal alignment context. This dataset is an updated version of the Everyday Activity sequences in the Aria Pilot Dataset. We updated the previous release using the most recent Machine Perception Services provided by Project Aria team and demonstrate a few research applications enabled by the machine perception data. We have also updated the open source Project Aria Tools to be compatible with this dataset and have provided a few research application examples. We believe AEA can serve as useful resources to enable researchers exploring always-on multimodal contextual AI research.

\section*{Acknowledgement}

The Aria Everyday Activities dataset is a collective effort from many contributors in Meta Reality Labs Research. The dataset's success depends on all the work and achievements of the entire Project Aria team and Machine Perception Services. We sincerely thank all the team members and partners who made this effort possible. In addition, we thank David Bui, Thomas Soares, Michael Loudon and Madalyn Bowen who participate in the data collection and perform the quality analysis for this dataset. 

{\small
\bibliographystyle{ieee_fullname}
\bibliography{egbib}

\begin{thebibliography}{10}\itemsep=-1pt

\bibitem{ego_procel}
Siddhant Bansal, Chetan Arora, and C.V. Jawahar.
\newblock My view is the best view: Procedure learning from egocentric videos.
\newblock In {\em European Conference on Computer Vision (ECCV)}, 2022.

\bibitem{dino}
Mathilde Caron, Hugo Touvron, Ishan Misra, Herv\'e J\'egou, Julien Mairal, Piotr Bojanowski, and Armand Joulin.
\newblock Emerging properties in self-supervised vision transformers.
\newblock In {\em Proceedings of the International Conference on Computer Vision (ICCV)}, 2021.

\bibitem{argoverse}
Ming-Fang Chang, John Lambert, Patsorn Sangkloy, Jagjeet Singh, Slawomir Bak, Andrew Hartnett, De Wang, Peter Carr, Simon Lucey, Deva Ramanan, and James Hays.
\newblock Argoverse: 3d tracking and forecasting with rich maps, 2019.

\bibitem{seamless2023}
Seamless Communication, Loïc Barrault, Yu-An Chung, Mariano~Coria Meglioli, David Dale, Ning Dong, Mark Duppenthaler, Paul-Ambroise Duquenne, Brian Ellis, Hady Elsahar, Justin Haaheim, John Hoffman, Min-Jae Hwang, Hirofumi Inaguma, Christopher Klaiber, Ilia Kulikov, Pengwei Li, Daniel Licht, Jean Maillard, Ruslan Mavlyutov, Alice Rakotoarison, Kaushik~Ram Sadagopan, Abinesh Ramakrishnan, Tuan Tran, Guillaume Wenzek, Yilin Yang, Ethan Ye, Ivan Evtimov, Pierre Fernandez, Cynthia Gao, Prangthip Hansanti, Elahe Kalbassi, Amanda Kallet, Artyom Kozhevnikov, Gabriel~Mejia Gonzalez, Robin~San Roman, Christophe Touret, Corinne Wong, Carleigh Wood, Bokai Yu, Pierre Andrews, Can Balioglu, Peng-Jen Chen, Marta~R. Costa-jussà, Maha Elbayad, Hongyu Gong, Francisco Guzmán, Kevin Heffernan, Somya Jain, Justine Kao, Ann Lee, Xutai Ma, Alex Mourachko, Benjamin Peloquin, Juan Pino, Sravya Popuri, Christophe Ropers, Safiyyah Saleem, Holger Schwenk, Anna Sun, Paden Tomasello, Changhan Wang, Jeff Wang, Skyler Wang, and Mary
  Williamson.
\newblock Seamless: Multilingual expressive and streaming speech translation, 2023.

\bibitem{epickitchens_scaling}
Dima Damen, Hazel Doughty, Giovanni~Maria Farinella, Sanja Fidler, Antonino Furnari, Evangelos Kazakos, Davide Moltisanti, Jonathan Munro, Toby Perrett, Will Price, and Michael Wray.
\newblock Scaling egocentric vision: The epic-kitchens dataset.
\newblock In {\em European Conference on Computer Vision (ECCV)}, 2018.

\bibitem{epickitchens100}
Dima Damen, Hazel Doughty, Giovanni~Maria Farinella, Antonino Furnari, Jian Ma, Evangelos Kazakos, Davide Moltisanti, Jonathan Munro, Toby Perrett, Will Price, and Michael Wray.
\newblock Rescaling egocentric vision: Collection, pipeline and challenges for epic-kitchens-100.
\newblock {\em International Journal of Computer Vision (IJCV)}, 130:33–55, 2022.

\bibitem{project_aria}
Jakob Engel, Kiran Somasundaram, Michael Goesele, Albert Sun, Alexander Gamino, Andrew Turner, Arjang Talattof, Arnie Yuan, Bilal Souti, Brighid Meredith, Cheng Peng, Chris Sweeney, Cole Wilson, Dan Barnes, Daniel DeTone, David Caruso, Derek Valleroy, Dinesh Ginjupalli, Duncan Frost, Edward Miller, Elias Mueggler, Evgeniy Oleinik, Fan Zhang, Guruprasad Somasundaram, Gustavo Solaira, Harry Lanaras, Henry Howard-Jenkins, Huixuan Tang, Hyo~Jin Kim, Jaime Rivera, Ji Luo, Jing Dong, Julian Straub, Kevin Bailey, Kevin Eckenhoff, Lingni Ma, Luis Pesqueira, Mark Schwesinger, Maurizio Monge, Nan Yang, Nick Charron, Nikhil Raina, Omkar Parkhi, Peter Borschowa, Pierre Moulon, Prince Gupta, Raul Mur-Artal, Robbie Pennington, Sachin Kulkarni, Sagar Miglani, Santosh Gondi, Saransh Solanki, Sean Diener, Shangyi Cheng, Simon Green, Steve Saarinen, Suvam Patra, Tassos Mourikis, Thomas Whelan, Tripti Singh, Vasileios Balntas, Vijay Baiyya, Wilson Dreewes, Xiaqing Pan, Yang Lou, Yipu Zhao, Yusuf Mansour, Yuyang Zou, Zhaoyang
  Lv, Zijian Wang, Mingfei Yan, Carl Ren, Renzo~De Nardi, and Richard Newcombe.
\newblock Project aria: A new tool for egocentric multi-modal ai research, 2023.

\bibitem{kitti}
Andreas Geiger, Philip Lenz, Christoph Stiller, and Raquel Urtasun.
\newblock Vision meets robotics: The kitti dataset.
\newblock {\em International Journal of Robotics Research (IJRR)}, 2013.

\bibitem{ego4d}
Kristen Grauman, Andrew Westbury, Eugene Byrne, Zachary Chavis, Antonino Furnari, Rohit Girdhar, Jackson Hamburger, Hao Jiang, Miao Liu, Xingyu Liu, Miguel Martin, Tushar Nagarajan, Ilija Radosavovic, Santhosh~Kumar Ramakrishnan, Fiona Ryan, Jayant Sharma, Michael Wray, Mengmeng Xu, Eric~Zhongcong Xu, Chen Zhao, Siddhant Bansal, Dhruv Batra, Vincent Cartillier, Sean Crane, Tien Do, Morrie Doulaty, Akshay Erapalli, Christoph Feichtenhofer, Adriano Fragomeni, Qichen Fu, Abrham Gebreselasie, Cristina Gonzalez, James Hillis, Xuhua Huang, Yifei Huang, Wenqi Jia, Weslie Khoo, Jachym Kolar, Satwik Kottur, Anurag Kumar, Federico Landini, Chao Li, Yanghao Li, Zhenqiang Li, Karttikeya Mangalam, Raghava Modhugu, Jonathan Munro, Tullie Murrell, Takumi Nishiyasu, Will Price, Paola~Ruiz Puentes, Merey Ramazanova, Leda Sari, Kiran Somasundaram, Audrey Southerland, Yusuke Sugano, Ruijie Tao, Minh Vo, Yuchen Wang, Xindi Wu, Takuma Yagi, Ziwei Zhao, Yunyi Zhu, Pablo Arbelaez, David Crandall, Dima Damen, Giovanni~Maria
  Farinella, Christian Fuegen, Bernard Ghanem, Vamsi~Krishna Ithapu, C.~V. Jawahar, Hanbyul Joo, Kris Kitani, Haizhou Li, Richard Newcombe, Aude Oliva, Hyun~Soo Park, James~M. Rehg, Yoichi Sato, Jianbo Shi, Mike~Zheng Shou, Antonio Torralba, Lorenzo Torresani, Mingfei Yan, and Jitendra Malik.
\newblock Ego4d: Around the world in 3,000 hours of egocentric video, 2022.

\bibitem{egoexo4d}
Kristen Grauman, Andrew Westbury, Lorenzo Torresani, Kris Kitani, Jitendra Malik, Triantafyllos Afouras, Kumar Ashutosh, Vijay Baiyya, Siddhant Bansal, Bikram Boote, Eugene Byrne, Zach Chavis, Joya Chen, Feng Cheng, Fu-Jen Chu, Sean Crane, Avijit Dasgupta, Jing Dong, Maria Escobar, Cristhian Forigua, Abrham Gebreselasie, Sanjay Haresh, Jing Huang, Md~Mohaiminul Islam, Suyog Jain, Rawal Khirodkar, Devansh Kukreja, Kevin~J Liang, Jia-Wei Liu, Sagnik Majumder, Yongsen Mao, Miguel Martin, Effrosyni Mavroudi, Tushar Nagarajan, Francesco Ragusa, Santhosh~Kumar Ramakrishnan, Luigi Seminara, Arjun Somayazulu, Yale Song, Shan Su, Zihui Xue, Edward Zhang, Jinxu Zhang, Angela Castillo, Changan Chen, Xinzhu Fu, Ryosuke Furuta, Cristina Gonzalez, Prince Gupta, Jiabo Hu, Yifei Huang, Yiming Huang, Weslie Khoo, Anush Kumar, Robert Kuo, Sach Lakhavani, Miao Liu, Mi Luo, Zhengyi Luo, Brighid Meredith, Austin Miller, Oluwatumininu Oguntola, Xiaqing Pan, Penny Peng, Shraman Pramanick, Merey Ramazanova, Fiona Ryan, Wei Shan,
  Kiran Somasundaram, Chenan Song, Audrey Southerland, Masatoshi Tateno, Huiyu Wang, Yuchen Wang, Takuma Yagi, Mingfei Yan, Xitong Yang, Zecheng Yu, Shengxin~Cindy Zha, Chen Zhao, Ziwei Zhao, Zhifan Zhu, Jeff Zhuo, Pablo Arbelaez, Gedas Bertasius, David Crandall, Dima Damen, Jakob Engel, Giovanni~Maria Farinella, Antonino Furnari, Bernard Ghanem, Judy Hoffman, C.~V. Jawahar, Richard Newcombe, Hyun~Soo Park, James~M. Rehg, Yoichi Sato, Manolis Savva, Jianbo Shi, Mike~Zheng Shou, and Michael Wray.
\newblock Ego-exo4d: Understanding skilled human activity from first- and third-person perspectives, 2023.

\bibitem{gaussian_splatting}
Bernhard Kerbl, Georgios Kopanas, Thomas Leimk{\"u}hler, and George Drettakis.
\newblock 3d gaussian splatting for real-time radiance field rendering.
\newblock {\em ACM Transactions on Graphics}, 42(4), July 2023.

\bibitem{sam}
Alexander Kirillov, Eric Mintun, Nikhila Ravi, Hanzi Mao, Chloe Rolland, Laura Gustafson, Tete Xiao, Spencer Whitehead, Alexander~C. Berg, Wan-Yen Lo, Piotr Doll{\'a}r, and Ross Girshick.
\newblock Segment anything.
\newblock {\em arXiv:2304.02643}, 2023.

\bibitem{egtea}
Yin Li, Miao Liu, and James~M. Rehg.
\newblock In the eye of beholder: Joint learning of gaze and actions in first person video.
\newblock In {\em Proceedings of the European Conference on Computer Vision (ECCV)}, September 2018.

\bibitem{gtea}
Yin Li, Zhefan Ye, and James~M Rehg.
\newblock Delving into egocentric actions.
\newblock In {\em Proceedings of the IEEE conference on computer vision and pattern recognition}, pages 287--295, 2015.

\bibitem{llava}
Haotian Liu, Chunyuan Li, Qingyang Wu, and Yong~Jae Lee.
\newblock Visual instruction tuning, 2023.

\bibitem{grounding_dino}
Shilong Liu, Zhaoyang Zeng, Tianhe Ren, Feng Li, Hao Zhang, Jie Yang, Chunyuan Li, Jianwei Yang, Hang Su, Jun Zhu, et~al.
\newblock Grounding dino: Marrying dino with grounded pre-training for open-set object detection.
\newblock {\em arXiv preprint arXiv:2303.05499}, 2023.

\bibitem{aria_pilot_dataset}
Zhaoyang Lv, Edward Miller, Jeff Meissner, Luis Pesqueira, Chris Sweeney, Jing Dong, Lingni Ma, Pratik Patel, Pierre Moulon, Kiran Somasundaram, Omkar Parkhi, Yuyang Zou, Nikhil Raina, Steve Saarinen, Yusuf~M Mansour, Po-Kang Huang, Zijian Wang, Anton Troynikov, Raul~Mur Artal, Daniel DeTone, Daniel Barnes, Elizabeth Argall, Andrey Lobanovskiy, David~Jaeyun Kim, Philippe Bouttefroy, Julian Straub, Jakob~Julian Engel, Prince Gupta, Mingfei Yan, Renzo~De Nardi, and Richard Newcombe.
\newblock Aria pilot dataset.
\newblock \url{https://about.facebook.com/realitylabs/projectaria/datasets}, 2022.

\bibitem{gpt4}
OpenAI, :, Josh Achiam, Steven Adler, Sandhini Agarwal, Lama Ahmad, Ilge Akkaya, Florencia~Leoni Aleman, Diogo Almeida, Janko Altenschmidt, Sam Altman, Shyamal Anadkat, Red Avila, Igor Babuschkin, Suchir Balaji, Valerie Balcom, Paul Baltescu, Haiming Bao, Mo Bavarian, Jeff Belgum, Irwan Bello, Jake Berdine, Gabriel Bernadett-Shapiro, Christopher Berner, Lenny Bogdonoff, Oleg Boiko, Madelaine Boyd, Anna-Luisa Brakman, Greg Brockman, Tim Brooks, Miles Brundage, Kevin Button, Trevor Cai, Rosie Campbell, Andrew Cann, Brittany Carey, Chelsea Carlson, Rory Carmichael, Brooke Chan, Che Chang, Fotis Chantzis, Derek Chen, Sully Chen, Ruby Chen, Jason Chen, Mark Chen, Ben Chess, Chester Cho, Casey Chu, Hyung~Won Chung, Dave Cummings, Jeremiah Currier, Yunxing Dai, Cory Decareaux, Thomas Degry, Noah Deutsch, Damien Deville, Arka Dhar, David Dohan, Steve Dowling, Sheila Dunning, Adrien Ecoffet, Atty Eleti, Tyna Eloundou, David Farhi, Liam Fedus, Niko Felix, Simón~Posada Fishman, Juston Forte, Isabella Fulford, Leo Gao,
  Elie Georges, Christian Gibson, Vik Goel, Tarun Gogineni, Gabriel Goh, Rapha Gontijo-Lopes, Jonathan Gordon, Morgan Grafstein, Scott Gray, Ryan Greene, Joshua Gross, Shixiang~Shane Gu, Yufei Guo, Chris Hallacy, Jesse Han, Jeff Harris, Yuchen He, Mike Heaton, Johannes Heidecke, Chris Hesse, Alan Hickey, Wade Hickey, Peter Hoeschele, Brandon Houghton, Kenny Hsu, Shengli Hu, Xin Hu, Joost Huizinga, Shantanu Jain, Shawn Jain, Joanne Jang, Angela Jiang, Roger Jiang, Haozhun Jin, Denny Jin, Shino Jomoto, Billie Jonn, Heewoo Jun, Tomer Kaftan, Łukasz Kaiser, Ali Kamali, Ingmar Kanitscheider, Nitish~Shirish Keskar, Tabarak Khan, Logan Kilpatrick, Jong~Wook Kim, Christina Kim, Yongjik Kim, Hendrik Kirchner, Jamie Kiros, Matt Knight, Daniel Kokotajlo, Łukasz Kondraciuk, Andrew Kondrich, Aris Konstantinidis, Kyle Kosic, Gretchen Krueger, Vishal Kuo, Michael Lampe, Ikai Lan, Teddy Lee, Jan Leike, Jade Leung, Daniel Levy, Chak~Ming Li, Rachel Lim, Molly Lin, Stephanie Lin, Mateusz Litwin, Theresa Lopez, Ryan Lowe,
  Patricia Lue, Anna Makanju, Kim Malfacini, Sam Manning, Todor Markov, Yaniv Markovski, Bianca Martin, Katie Mayer, Andrew Mayne, Bob McGrew, Scott~Mayer McKinney, Christine McLeavey, Paul McMillan, Jake McNeil, David Medina, Aalok Mehta, Jacob Menick, Luke Metz, Andrey Mishchenko, Pamela Mishkin, Vinnie Monaco, Evan Morikawa, Daniel Mossing, Tong Mu, Mira Murati, Oleg Murk, David Mély, Ashvin Nair, Reiichiro Nakano, Rajeev Nayak, Arvind Neelakantan, Richard Ngo, Hyeonwoo Noh, Long Ouyang, Cullen O'Keefe, Jakub Pachocki, Alex Paino, Joe Palermo, Ashley Pantuliano, Giambattista Parascandolo, Joel Parish, Emy Parparita, Alex Passos, Mikhail Pavlov, Andrew Peng, Adam Perelman, Filipe de Avila Belbute~Peres, Michael Petrov, Henrique~Ponde de Oliveira~Pinto, Michael, Pokorny, Michelle Pokrass, Vitchyr Pong, Tolly Powell, Alethea Power, Boris Power, Elizabeth Proehl, Raul Puri, Alec Radford, Jack Rae, Aditya Ramesh, Cameron Raymond, Francis Real, Kendra Rimbach, Carl Ross, Bob Rotsted, Henri Roussez, Nick Ryder,
  Mario Saltarelli, Ted Sanders, Shibani Santurkar, Girish Sastry, Heather Schmidt, David Schnurr, John Schulman, Daniel Selsam, Kyla Sheppard, Toki Sherbakov, Jessica Shieh, Sarah Shoker, Pranav Shyam, Szymon Sidor, Eric Sigler, Maddie Simens, Jordan Sitkin, Katarina Slama, Ian Sohl, Benjamin Sokolowsky, Yang Song, Natalie Staudacher, Felipe~Petroski Such, Natalie Summers, Ilya Sutskever, Jie Tang, Nikolas Tezak, Madeleine Thompson, Phil Tillet, Amin Tootoonchian, Elizabeth Tseng, Preston Tuggle, Nick Turley, Jerry Tworek, Juan Felipe~Cerón Uribe, Andrea Vallone, Arun Vijayvergiya, Chelsea Voss, Carroll Wainwright, Justin~Jay Wang, Alvin Wang, Ben Wang, Jonathan Ward, Jason Wei, CJ Weinmann, Akila Welihinda, Peter Welinder, Jiayi Weng, Lilian Weng, Matt Wiethoff, Dave Willner, Clemens Winter, Samuel Wolrich, Hannah Wong, Lauren Workman, Sherwin Wu, Jeff Wu, Michael Wu, Kai Xiao, Tao Xu, Sarah Yoo, Kevin Yu, Qiming Yuan, Wojciech Zaremba, Rowan Zellers, Chong Zhang, Marvin Zhang, Shengjia Zhao, Tianhao
  Zheng, Juntang Zhuang, William Zhuk, and Barret Zoph.
\newblock Gpt-4 technical report, 2023.

\bibitem{adt}
Xiaqing Pan, Nicholas Charron, Yongqian Yang, Scott Peters, Thomas Whelan, Chen Kong, Omkar Parkhi, Richard Newcombe, and Carl~Yuheng Ren.
\newblock Aria digital twin: A new benchmark dataset for egocentric 3d machine perception, 2023.

\bibitem{clip}
Alec Radford, Jong~Wook Kim, Chris Hallacy, Aditya Ramesh, Gabriel Goh, Sandhini Agarwal, Girish Sastry, Amanda Askell, Pamela Mishkin, Jack Clark, Gretchen Krueger, and Ilya Sutskever.
\newblock Learning transferable visual models from natural language supervision, 2021.

\bibitem{whisper}
Alec Radford, Jong~Wook Kim, Tao Xu, Greg Brockman, Christine McLeavey, and Ilya Sutskever.
\newblock Robust speech recognition via large-scale weak supervision.
\newblock In {\em International Conference on Machine Learning}, pages 28492--28518. PMLR, 2023.

\bibitem{schoenberger2016sfm}
Johannes~Lutz Sch\"{o}nberger and Jan-Michael Frahm.
\newblock Structure-from-motion revisited.
\newblock In {\em Conference on Computer Vision and Pattern Recognition (CVPR)}, 2016.

\bibitem{assembly101}
F. Sener, D. Chatterjee, D. Shelepov, K. He, D. Singhania, R. Wang, and A. Yao.
\newblock Assembly101: A large-scale multi-view video dataset for understanding procedural activities.
\newblock {\em CVPR 2022}, 2022.

\bibitem{waymo}
Pei Sun, Henrik Kretzschmar, Xerxes Dotiwalla, Aurelien Chouard, Vijaysai Patnaik, Paul Tsui, James Guo, Yin Zhou, Yuning Chai, Benjamin Caine, Vijay Vasudevan, Wei Han, Jiquan Ngiam, Hang Zhao, Aleksei Timofeev, Scott Ettinger, Maxim Krivokon, Amy Gao, Aditya Joshi, Yu Zhang, Jonathon Shlens, Zhifeng Chen, and Dragomir Anguelov.
\newblock Scalability in perception for autonomous driving: Waymo open dataset.
\newblock In {\em Proceedings of the IEEE/CVF Conference on Computer Vision and Pattern Recognition (CVPR)}, June 2020.

\bibitem{nerfstudio}
Matthew Tancik, Ethan Weber, Evonne Ng, Ruilong Li, Brent Yi, Justin Kerr, Terrance Wang, Alexander Kristoffersen, Jake Austin, Kamyar Salahi, Abhik Ahuja, David McAllister, and Angjoo Kanazawa.
\newblock Nerfstudio: A modular framework for neural radiance field development.
\newblock In {\em ACM SIGGRAPH 2023 Conference Proceedings}, SIGGRAPH '23, 2023.

\bibitem{gemini}
Gemini Team.
\newblock Gemini: A family of highly capable multimodal models, 2023.

\bibitem{epic_fields}
Vadim Tschernezki, Ahmad Darkhalil, Zhifan Zhu, David Fouhey, Iro Larina, Diane Larlus, Dima Damen, and Andrea Vedaldi.
\newblock {EPIC Fields}: {M}arrying {3D} {G}eometry and {V}ideo {U}nderstanding.
\newblock In {\em Proceedings of the Neural Information Processing Systems (NeurIPS)}, 2023.

\bibitem{holo_assist}
Xin Wang, Taein Kwon, Mahdi Rad, Bowen Pan, Ishani Chakraborty, Sean Andrist, Dan Bohus, Ashley Feniello, Bugra Tekin, Felipe~Vieira Frujeri, Neel Joshi, and Marc Pollefeys.
\newblock Holoassist: an egocentric human interaction dataset for interactive ai assistants in the real world.
\newblock In {\em Proceedings of the IEEE/CVF International Conference on Computer Vision (ICCV)}, pages 20270--20281, October 2023.

\bibitem{efficientsam}
Yunyang Xiong, Bala Varadarajan, Lemeng Wu, Xiaoyu Xiang, Fanyi Xiao, Chenchen Zhu, Xiaoliang Dai, Dilin Wang, Fei Sun, Forrest Iandola, Raghuraman Krishnamoorthi, and Vikas Chandra.
\newblock Efficientsam: Leveraged masked image pretraining for efficient segment anything.
\newblock 2023.

\end{thebibliography}
}

\appendix

\section{Project Aria Raw Sensor Data Description}
\label{sec:raw_sensor}

Project Aria devices integrate a variety of sensors that record egocentric multimodal data that can be configured optionally at run-time by users. For a more comprehensive description of Project Aria devices, readers can refer to the device description in Project Aria white paper \cite{project_aria}). 

To fully empower multi-modal research in everyday scenarios, we chose a recording configuration that provided a rich suite of sensors appropriate for an indoor setting.  The sensors used to generate raw data were: 

\paragraph{RGB Camera:} This is a rolling shuttle camera with $110^\circ$ horizontal field of view (HFoV). We captured the RGB stream at 20 fps with a resolution of 1408x1408 pixels.

\paragraph{Monochrome Scene / SLAM Cameras:} These cameras are global-shutter monochrome cameras with $150^\circ$ HFoV, placed on left and right side of the glasses. These two cameras are primarily useful for visual SLAM and can also be helpful for hand tracking. We recorded each monochrome video stream at 10 fps with a resolution of 640x480 pixels. 

\paragraph{Eye Tracking Cameras:} There are two monochrome inward facing cameras with IR illumination used for eye tracking purposes. They recorded at 10 fps with a resolution of 320x240 pixels and $80^\circ$ diagonal field of view (DFoV). 

\paragraph{IMUs:} There are two inertial measurements units (IMU) on each side of the glasses. The left IMU runs at 800 Hz with a saturation limit of 4g (accelerometer) and $500^\circ /s$ (gyroscope). The right IMU samples at 1000 Hz with a saturation limit of 8g (accelerator) and $1000^\circ/s$ (gyroscope).

\paragraph{Microphone:} There are 7 microphones in the glasses, five face the front plus one on each side. We captured spatial audio with 7 channels using a sample rate of 48kHz. 

\paragraph{Other sensors:} A magnetometer measured the ambient magnetic field
with a resolution of $0.1 µT$ at a sample rate of 10 Hz. A barometer sensor
captured local air pressure and temperature at a resolution of $0.66 Pa$ and $0.005° C$, respectively at a sample rate of 50 Hz. We did not turn on GPS or WiFi sensors during any of the recordings. 

\section{Machine Perception Services (MPS)}
\label{sec:mps}

In addition to raw sensor data, we've provided a range of machine perception capabilities that further empower downstream applications that require reliable 3D and eyegaze information derived from the sensors. We acquire the location and eye gaze information using the Machine Perception Services (MPS) provided by Project Aria, which has superior accuracy and robustness compared existing off-the-shelf open source solutions. For speech transcription, we used an in-house speech recognition model. Please refer to \cite{project_aria} for more details about Machine Perception Services.

\paragraph{Open-loop trajectory:} For each of the recording, we provided a high frequency (1 kHz) 6DoF trajectory from the real-time odometry estimation.
This trajectory is suitable to be used to mimic any real-time applications that require on-device 6DoF pose information. 

\paragraph{Close-loop trajectory:} 
This is also a high frequency (1 kHz) 6 DoF trajectory, but acquired via post-processing. Unlike open-loop trajectory, this trajectory shares the same reference frame from all other sequences recorded at the same location. This ensures we can align all the trajectories from different recordings by multiple wearers doing multiple activities in the same location. 

\paragraph{Semi-dense Points:} 
In addition to the close-loop trajectory, we've provided semi-dense tracks and point cloud using the same global frame of reference, which is a partial reconstruction of the static portion of the environment.  Fig.~\ref{fig:multi_user_activity} shows the visualized point cloud together with closed-loop trajectory from all different activities performed by one user in the same location during a day. 

\paragraph{Eye-gaze tracking:} We've provided calculated 3D eye gaze from the Project Aria device eye tracking
cameras. This is represented as a per-frame 3D ray anchored to the central pupil frame. Using device calibration, we can also project the ray to each 2D camera and get an eye gaze location using an estimated depth along each ray. The gaze information is an important indicator of wearer's attention, which can be helpful for a variety of contextual AI applications. 

\section{Activity Scripts in Data Collection}
\label{sec:activity_scripts}

We provided five main recordings scripts as guidance in the data collection. In each location, the wearers will perform guided activities follow all the high-level scripts. Each script contained multiple scenarios that told a story about people going through their day. 

\begin{itemize}
    \item Script 1: A lazy morning before the party. – Single wearer
    \item Script 2: Catch up and have some fun. – Two wearers 
    \item Script 3: What do you want for dinner? – Two wearers
    \item Script 4: Easy like Sunday morning. – Single wearer
    \item Script 5: Get home then get going. – Single wearer
\end{itemize}

Each script is connected by multiple sequential activities in a day. We provide the guidance activity name paired with each script ID and sequence ID in Tab.\ref{tab:script_description}. 

We provide downloading scripts for readers to download the data in one location, which contain all available sequences in all scripts. It is worth noting that not all activities are available to be downloaded for each of the script in every location. We have done a rigorous quality analysis on the collected recordings, which have filtered out certain recordings that do not pass our quality analysis.   

We further categorize the activities into a few common taxonomies and provide the list to identify where scenarios are used in scripts. The Wearer numbers indicate how many wearers were in a sequence. We provide the details of the activity to scripts mapping in Tab.\ref{tab:activity_scripts}.

\begin{table}[h]
\centering
\begin{tabular}{ccl}
\toprule
\centering
\textbf{Script ID} & \textbf{Sequence ID} & \textbf{Open World Activities} \\
\hline
1 & 1 & Health activity \\
1 & 2 & Watch TV \\
1 & 3 & Clean the place \\
1 & 4 & Room Decoration  \\
1 & 5 & Video game break \\
1 & 6 & Food and accomodations \\
1 & 7 & Read and wait \\
2 & 1 & Friends arrives at home \\
2 & 2 & The home tour \\
2 & 3 & Caffeination \\
2 & 4 & Grab cream, sugar and snacks \\
2 & 5 & Play a board game \\
2 & 6 & Clean a spill \\
2 & 7 & Share some vides \\
2 & 8 & Time to go \\
3 & 1 & Texting and Reading \\
3 & 2 & Talk about the day \\
3 & 3 & Cooking \\
3 & 4 & Set the table and eat \\
3 & 5 & Clean up \\
4 & 1 & Waking up \\
4 & 2 & Perk up with coffee \\
4 & 3 & Morning exercise \\
4 & 4 & Prepare breakfast \\
4 & 5 & Eating \\
4 & 6 & Clean up \\
4 & 7 & Relaxing \\
5 & 1 & Get home \\
5 & 2 & Wash the clothes \\
5 & 3 & Straighten up \\
5 & 4 & Dry your clothes \\
5 & 5 & Catch up on what happened \\
5 & 6 & Get and fold the laundry \\
5 & 7 & Check the food provisions \\
\bottomrule
\end{tabular}
\caption{The activities mapped to each sequence in the recording scripts.}
\label{tab:script_description}
\end{table}

\begin{table*}[t]
\centering
\begin{tabular}{llc}
\hline
\centering
\textbf{Task} & \textbf{Script ID} & \textbf{Number of Wearers}\\ \toprule
Making coffee & Script 2: Caffeination & 2 \\
Making coffee & Script 4: Perk up  & 1 \\ 
\midrule
Prepare snacks & Script 1: Set out food and drink & 1 \\
Prepare snacks & Script 2: Grab the cream, sugar, and some snacks & 2 \\ 
\midrule
Cooking & Script 3: Cooking & 2 \\
Cooking & Script 4: Breaking the Fast & 1 \\
Cooking & Script 5: Check the provisions & 1 \\ 
\midrule 
Clean the place & Script 1: Clean the place & 1 \\ 
Clean the place & Script 2: Clean up spilled coffee & 2 \\
Clean the place & Script 2: Time to go & 2 \\
Clean the place & Script 3: Clean up & 2 \\
Clean the place & Script 4: Cleaning up & 1 \\ 
\midrule
Dining & Script 3: Set the table & 2 \\
Dining & Script 4: Eating & 1 \\ 
\midrule
Organization and laundry & Script 1: Put up decorations & 1 \\
Organization and laundry & Script 5: Get home & 1 \\
Organization and laundry & Script 5: Wash clothes & 1 \\
Organization and laundry & Script 5: Straighten up & 1 \\
Organization and laundry & Script 5: Dry clothes & 1 \\
Organization and laundry & Script 5: Get and fold laundry & 1 \\ 
\midrule
Reading, games and exercise & Script 1: Get the blood pumping & 1 \\
Reading, games and exercise & Script 1: Watch a TV Show & 1 \\
Reading, games and exercise & Script 1: Video game break& 1 \\
Reading, games and exercise & Script 1: Read and wait & 1 \\
Reading, games and exercise & Script 4: Waking up & 1 \\
Reading, games and exercise & Script 4: Exercise & 1 \\
Reading, games and exercise & Script 4: Play console video game & 1 \\
Reading, games and exercise & Script 5: Catch up on email and social media  & 1 \\ 
\midrule
Touring the room & Script 2: Tour of the house & 2 \\ 
\midrule
Multi-person indoor activities & Script 2: Guest arrives & 2 \\
Multi-person indoor activities & Script 2: play a board game & 2 \\
Multi-person indoor activities & Script 2: Share videos & 2 \\
Multi-person indoor activities & Script 3: Arriving home & 2 \\ 
\midrule
Activities with indoor outdoor transitions & Script 2: Guest arrives & 2 \\
Activities with indoor outdoor transitions & Script 3: Arriving home & 2 \\
Activities with indoor outdoor transitions & Script 5: Get home & 1 \\
\bottomrule
\end{tabular}
\caption{The details of all activities mapped to each recording script.}
\label{tab:activity_scripts}
\end{table*}

\end{document}